\documentclass[showkeys]{revtex4}
\usepackage{graphicx}
\usepackage{fancyhdr}
\usepackage{amsmath,amsfonts,amssymb}
\usepackage{enumitem}

\newtheorem{definition}{Definition}
\newtheorem{proposition}{Proposition}

\usepackage{pgf,tikz}
\usetikzlibrary{arrows}

\usepackage[a4paper]{geometry}
\usepackage{color}
\definecolor{red}{rgb}{1,0,0}
\definecolor{gre}{rgb}{0,1,0}
\definecolor{blu}{rgb}{0.05, 0.11, 0.51}
\definecolor{white}{rgb}{1,1,1}

\newcommand{\sk}{\vskip0.3truecm}

\newcommand{\cc}[1]{}

\usepackage{hyperref}   
\hypersetup{
bookmarks=true,           
unicode=false,          
pdftoolbar=true,        
pdfmenubar=true,        
pdffitwindow=false,     
pdfstartview={FitH},    
pdftitle={Knowledge as Fruits of Ignorance : a probabilistic approach of our way of thinking },    
pdfauthor={Thomas Cailleteau aka Thomaths Quantix, Aur\'elien Barrau},     
pdfsubject={Entropy, Ignorance, Jaynes and Shannon},   
pdfcreator={Thomas Cailleteau, Aur\'elien Barrau},   
pdfproducer={Producer}, 
pdfkeywords={entropy} {ignorance} {jaynes} {Shannon}, 
pdfnewwindow=true,      
breaklinks=true,       
urlcolor= blu,        
linkcolor= red,        
colorlinks=true,       
linkcolor=blu, 
citecolor=blu,        
filecolor=blu,      
urlcolor=blue           
}

\begin{document}


\title{Knowledge as Fruits of Ignorance :\\ A kind of global free energy principle of our way of thinking } 
\author{Thomas Cailleteau}

\affiliation{
Sant Job Skolaj-Lise,  42 Kerguestenen Straed, 56100 BroAnOriant, Breizh}
\date{6 Juin 2022}

\keywords{Entropy, Ignorance, Shannon, Jaynes, Bayes, Free Energy Principle}

\email{thomas.cailleteau@lpsc.in2p3.fr}

\begin{abstract}
In this second article, we show a simple use of the Ignorance as defined in a previous article \textit{Jaynes \& Shannon's Constrained Ignorance and Surprise}. By giving an example about the journey of a person, we believe to show some simple, obvious but mathematically encoded philosophical implications about how we could think, learn and memorize. In this basic model we will separate how we learn from Ignorance, and how we anticipate the world using Bayes formula, both should however be more entangled to best reflect reality. In fact, as we have seen after achieving this work, applying ignorance on the system constituting a person finally turns out to be the global approach of its local counterpart on systems like neurons, cells and other complex probabilistic systems, described using the free energy principle, a much more complexe and detailed approach. The aim of this article is therefore to show, as seen from a person, another aspect of the application of the free energy principle which represents the constrained Shannon's entropy, and leads to Bayes'formula. \sk
We show that, using only ignorance as a single quantity, and its minimisation as the main process, we can take into account his understandings, assertions, doubts and assumptions about how he perceives the world, by describing them mathematically. As in the following we will be assertive and provocative on purpose, any comments are welcomed and would be appreciated. 
\end{abstract}

\maketitle

\section{Introduction}
This work, grounded first in Edwin T. Jaynes' book \textit{Probability Theory: The Logic of Science} \cite{Jaynesbook}, relies on a previous article \cite{MyIgnorance} \textit{Jaynes \& Shannon's Constrained Ignorance and Surprise} about, what we call, Ignorance. As shown in \cite{MyIgnorance}, from a probability tree and a simple mathematical definition, using constraints, of Ignorance (related to knowledge), one is able to derive Shannon's Entropy as in Jaynes' maximum entropy principle by maximizing/minimizing it. Using Shannon's Entropy with constraints was in fact used before, in different approaches but with much more mathematical details : see for instance \cite{GiffinCaticha} or \cite{BanavarMaritan}, but also \cite{KFriston}, \cite{Sakth} for its use in the framework of the  Free Energy Principle, \cite{FEP} for a recent review and references within. At the time we were redacting this article, we were not aware about all of this, but results are really similar, even if the way to achieve it is different. In fact, here, applying Ignorance on the system constituting a person finally seems to turn out to be the global approach of its local counterpart on systems like neurons, cells, ... and other complex probabilistic systems, described using the \textit{Free Energy Principle}. \sk

In our opinion, the framework of Ignorance leads to a more understandable way of seeing the entropy and its consequences, as we start only from a tree of possibilities, and what we know.  Moreover, as it seemed to be surprising that Bayes formula appeared in the calculations in the articles previously cited, we think that here Bayes formula is clearly seen, as it should be, using a probability tree.\sk 

In this article, we will follow the footsteps of an Ignorant person called \textit{Howard}, and study the stages of his learning. Taking some liberties, we assume that Howard has amnesia (in order to start from a simili zero blank state) but has also some primary knowledge that he is not aware of (he can \textit{a priori} walk, move, listen, read, understand and speak). 
\sk 

During this journey, we will discuss a mathematical construction with some possible applications on the global (!) porcessus of learning and thinking, leading to, what we think are, philosophical consequences. We are also deliberately vague about what "things/stuffs" are, because we want to show mainly the learning process and its philosophical implications, without going into other ontological philosophical aspects which go deeper in the way of describing reality. Therefore, as it is not easy to describe what "knowledge / to know" means, we will use common sense and talk about it another time, with a more extended framework using what is shown here. However, about what concerns us here, in \cite{MyIgnorance}, we have decided to describe mathematically \textit{what we knows}, by taking the opposite view, that is, from the prism about \textit{what we ignore}, as we are mostly Ignorant, using probabilities : because of our memory, its uncertainties but also those of our perceptions, we are never confident in our (relative) knowledge, and therefore we should speak in terms of likelihood, plausibility, degree of confidence, ... in other words, in terms of probabilities about our situation.

\section{At first he knows nothing except that he doesn't know much}

We know what we know but don't know what we don't ... Let us assume that, in a corridor, walks Howard, depicted by a system $\mathcal{S}$ in a state ${Z}$ (this state depends on the path took by Howard in his life, his knowledges and choices that lead him, at this point in time and space, in this corridor). Therefore, all what he will learn will be through his perceptions and relatively to him. What can he say at first ? Something like \textit{"As I am and think, things exist"}, which can be translated in a proposition as 
\begin{itemize}
\item $E_0$ : "Things exist, like time, space and stuffs I can perceive " 
\end{itemize}

So, the \textbf{probability for Howard that $E_1$ is true} is .. 1 : $P(E_0 == True | Z ) = 1$, that is it is a certain event, in this particular case "even before" he has experimented it. We can then define a quantity, called Ignorance, such that

\begin{definition}
The \textbf{\textit{Ignorance $h$ of (this) knowledge}} is defined as
\begin{equation} \label{eq:ignorance_nulle_EO} 
h_0(E_0, Z) = P(E_0 == True | Z ) - 1 \,\,\,\, \rightarrow \,\,\,\, \ 0
\end{equation}
which, being a constraint on the system $\mathcal{S}$, is null by definition.
\end{definition}

In the following, except when needed, we will drop the "$==True $" as we evaluate the truthfulness of a proposition at a moment $t$ in space $\vec{x}$, in the system of coordinates $(x^\mu)$, $\mu \in \{0,1,...\}$, with the knowledge available at this point in Howard space-time. For these reasons, we say that knowledge are relative, or better, in French, as the word "knowledge" is too narrow to express all the subtilities of "Nature", \textit{la connaissance est relative mais le savoir est absolu}. That's also why in Eq.(\ref{eq:ignorance_nulle_EO}), we have set an arrow to express that, even if Howard "knows" that $E_0$ is true, it may be not a reality except for him. 
\sk

We can thus summarize his update of knowledge by the evolution $Z \rightarrow Z_0$ for his state of knowledge (omitting, as said before, many things because of his amnesia) at this point in spacetime, and illustrating it with the following weighted probability tree (here with only two branches due to the only assertion $E_0$) : 

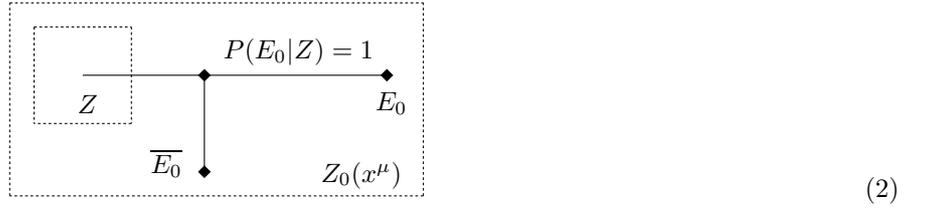
\begin{equation}
\begin{tikzpicture}[scale = 1.6]
\draw (6.62,-0.34)-- (5.12,-0.34);
\draw (5.2,0.04) node[anchor=north west] {$P(E_0 | Z ) = 1$};
\draw (6.45,-0.4) node[anchor=north west] {$E_0$};
\draw (6,-0.98) node[anchor=north west] {$Z_0(x^\mu)$};
\draw (5.12,-0.34)-- (5.12,-1.14);
\draw (4.6,-0.88) node[anchor=north west] {$\overline{E_0}$};
\draw (5.12,-0.34)-- (4.12,-0.34);
\draw (4,-0.43) node[anchor=north west] {$Z$};
\draw [dash pattern=on 1pt off 1pt] (3.72,0.06)-- (4.52,0.06);
\draw [dash pattern=on 1pt off 1pt] (4.52,0.06)-- (4.52,-0.74);
\draw [dash pattern=on 1pt off 1pt] (4.52,-0.74)-- (3.72,-0.74);
\draw [dash pattern=on 1pt off 1pt] (3.72,0.06)-- (3.72,-0.74);
\draw [dash pattern=on 1pt off 1pt] (3.52,0.26)-- (6.92,0.26);
\draw [dash pattern=on 1pt off 1pt] (6.92,-1.34)-- (6.92,0.26);
\draw [dash pattern=on 1pt off 1pt] (6.92,-1.34)-- (3.52,-1.34);
\draw [dash pattern=on 1pt off 1pt] (3.52,-1.34)-- (3.52,0.26);
\begin{scriptsize}
\fill [color=black] (6.62,-0.34) ++(-1.5pt,0 pt) -- ++(1.5pt,1.5pt)--++(1.5pt,-1.5pt)--++(-1.5pt,-1.5pt)--++(-1.5pt,1.5pt);
\fill [color=black] (5.12,-0.34) ++(-1.5pt,0 pt) -- ++(1.5pt,1.5pt)--++(1.5pt,-1.5pt)--++(-1.5pt,-1.5pt)--++(-1.5pt,1.5pt);
\fill [color=black] (5.12,-1.14) ++(-1.5pt,0 pt) -- ++(1.5pt,1.5pt)--++(1.5pt,-1.5pt)--++(-1.5pt,-1.5pt)--++(-1.5pt,1.5pt);
\end{scriptsize}
\end{tikzpicture}
\end{equation}

\section{Second piece of knowledge}
 
Walking down the corridor, he comes to a door with a frame beside it containing a piece of paper that says "Somewhere, there is a coin and a dice". Howard doesn't wonder about the door, nor about the fact that he can read and understand, but he does wonder about the things he has forgotten, the coin and the dice. In his mind, as he knows what means "there is", appear two new propositions : 
\begin{itemize}
\item $E_1$ : "A coin exists",
\item $E_2$ : "A dice exists",
\end{itemize} 

Reading the paper and having therefore some informations about what can exist, has lead Howard to update his knowledge such that his state is now $Z_0(x^\mu) \rightarrow Z_1(Z_0(x^\mu),x'{}^{\mu}) $, that is, in terms of probabilities 
\begin{equation}  
P({E_1 \cup E_2}|Z_1) + P(\overline{E_1 \cup E_2}|Z_1 ) \,\,=  \,\, P(E_1|Z_1) + P(E_2|Z_1) + P(\overline{E_1 \cup E_2}|Z_1) \,\, =  \,\, 1, 
\end{equation}
as $E_1$ and $E_2$ are independent. Indeed, the creation of the dice may depend on the creation of the coin, but the idea that it exists does not depend on the idea that the coin exists, as here, Howard has no other knowledge about the construction of the dice and the coin. $\overline{E_1 \cup E_2}$ represents everything that is not $E_1$ or $E_2$ and can therefore be anything. His Ignorance for what he knows about his situation, after this update, is now defined as 
\begin{equation}\label{eq:constraint2}
h_1[E_1,E_2,\overline{E_1 \cup E_2} | Z_0] =   P(E_1|Z_0) + P(E_2|Z_0) + P(\overline{E_1 \cup E_2}|Z_0) -1  = 0
\end{equation} 
and we can represent the evolution of his state of knowledge as 
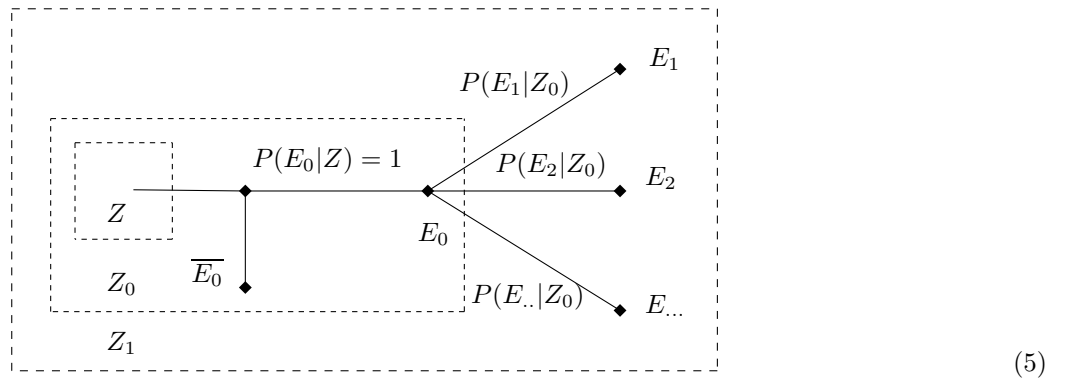
\begin{equation}\label{graph:stateatZ1}
\begin{tikzpicture}[scale = 1.6] 
\draw (6.92,-0.51)-- (5.42,-0.51);
\draw (5.4,-0.07) node[anchor=north west] {$P(E_0 | Z ) = 1$};
\draw (6.76,-0.7) node[anchor=north west] {$E_0$};
\draw (4.2,-1.11) node[anchor=north west] {$Z_0$};
\draw (5.42,-0.51)-- (5.42,-1.31);
\draw (4.9,-0.99) node[anchor=north west] {$\overline{E_0}$};
\draw (4.2,-0.54) node[anchor=north west] {$Z$};
\draw [line width=0.4pt,dash pattern=on 2pt off 2pt] (4.02,-0.11)-- (4.82,-0.11);
\draw [line width=0.4pt,dash pattern=on 2pt off 2pt] (4.82,-0.11)-- (4.82,-0.91);
\draw [line width=0.4pt,dash pattern=on 2pt off 2pt] (4.82,-0.91)-- (4.02,-0.91);
\draw [line width=0.4pt,dash pattern=on 2pt off 2pt] (4.02,-0.11)-- (4.02,-0.91);
\draw [dash pattern=on 2pt off 2pt] (3.82,0.09)-- (7.22,0.09);
\draw [dash pattern=on 2pt off 2pt] (7.22,-1.51)-- (7.22,0.09);
\draw [dash pattern=on 2pt off 2pt] (7.22,-1.51)-- (3.82,-1.51);
\draw [dash pattern=on 2pt off 2pt] (3.82,-1.51)-- (3.82,0.09);
\draw (6.92,-0.51)-- (8.5,0.5);
\draw (6.92,-0.51)-- (8.5,-0.51);
\draw (6.92,-0.51)-- (8.5,-1.5);
\draw (8.66,0.75) node[anchor=north west] {$E_1$};
\draw (8.63,-0.23) node[anchor=north west] {$E_2$};
\draw (8.63,-1.29) node[anchor=north west] {$E_{...}$};
\draw (7.1,0.56) node[anchor=north west] {$P(E_1 | Z_0 )$};
\draw (7.4,-0.11) node[anchor=north west] {$P(E_2 | Z_0)$};
\draw (7.2,-1.2) node[anchor=north west] {$P( {E_{..}} | Z_0)$};
\draw (5.42,-0.51)-- (4.5,-0.5);
\draw [dash pattern=on 3pt off 3pt] (3.5,1)-- (9.3,1);
\draw [dash pattern=on 3pt off 3pt] (9.3,1)-- (9.3,-2);
\draw [dash pattern=on 3pt off 3pt] (9.3,-2)-- (3.5,-2);
\draw [dash pattern=on 3pt off 3pt] (3.5,-2)-- (3.5,1);
\draw (4.2,-1.6) node[anchor=north west] {$Z_1$};
\begin{scriptsize}
\fill [color=black] (6.92,-0.51) ++(-1.5pt,0 pt) -- ++(1.5pt,1.5pt)--++(1.5pt,-1.5pt)--++(-1.5pt,-1.5pt)--++(-1.5pt,1.5pt);
\fill [color=black] (5.42,-0.51) ++(-1.5pt,0 pt) -- ++(1.5pt,1.5pt)--++(1.5pt,-1.5pt)--++(-1.5pt,-1.5pt)--++(-1.5pt,1.5pt);
\fill [color=black] (5.42,-1.31) ++(-1.5pt,0 pt) -- ++(1.5pt,1.5pt)--++(1.5pt,-1.5pt)--++(-1.5pt,-1.5pt)--++(-1.5pt,1.5pt);
\fill [color=black] (8.5,0.5) ++(-1.5pt,0 pt) -- ++(1.5pt,1.5pt)--++(1.5pt,-1.5pt)--++(-1.5pt,-1.5pt)--++(-1.5pt,1.5pt);
\fill [color=black] (8.5,-0.51) ++(-1.5pt,0 pt) -- ++(1.5pt,1.5pt)--++(1.5pt,-1.5pt)--++(-1.5pt,-1.5pt)--++(-1.5pt,1.5pt);
\fill [color=black] (8.5,-1.5) ++(-1.5pt,0 pt) -- ++(1.5pt,1.5pt)--++(1.5pt,-1.5pt)--++(-1.5pt,-1.5pt)--++(-1.5pt,1.5pt);
\end{scriptsize}
\end{tikzpicture}
\end{equation}
where we set $\overline{E_1\cup E_2} = E..$ for short. In \cite{MyIgnorance}, we have shown that, in this case, the update of knowledge leads us to define another kind of Ignorance, the one about things he does not know yet, if they will happen, are real or not, .. \textit{etc etc}, which is simply given by the sum of the different entropies : 
\begin{definition}
The \textbf{\textit{Ignorance $\bar{h}$ of what he does not know yet}} can naturally be described as the sum of the diverse entropies that exist due to each event : 
\begin{equation}
\bar{h}_1[E_1,E_2,E_{..} | Z_1(Z_0(x^\mu),x'{}^\mu)]  = - \displaystyle \,\, \sum_i \,\, \lambda_i(E_i, Z_1(Z_0(x^\mu),x'{}^\mu)) \,\, P(E_i |Z_0) \times \ln (\,\, P(E_i | Z_0)\,\, ),
\end{equation}
where $\lambda_i(E_i, Z_n,t,\vec{x})$, Lagrange multipliers, are coefficients we could assume not to be 0 or 1. In the following, it will be seen as only $\lambda$, as any other greek letters which have the same kind of properties.
\end{definition}
\begin{definition}
The (whole) Ignorance $H$ is given by the Ignorance $h$ of what is known, as constraints, therefore using Lagrange multipliers, and the Ignorance $\bar{h}$ of what remains to be known, which can be calculated as it consists of entropies.
\end{definition}
For Howard, at this point, following the probability tree and the fact that $Z_1(Z_0(x^\mu),x'{}^\mu)$, his Ignorance is thus 
\begin{eqnarray} \label{ex:fundamental_ignorance}
& & H[E_0, E_1,E_2,\overline{E_1 \cup E_2}, \lambda, \mu | Z_1(Z_0(x^\mu),x'{}^\mu)] \nonumber \\ &=& \mu_0 \times  h_0[E_0, Z_0] + \mu_1 \times h_1[E_1,E_2,\overline{E_1 \cup E_2}, Z_0] + \bar{h}_1[E_1,E_2,\overline{E_1 \cup E_2}, Z_0] \nonumber \\
&=& \displaystyle \mu_0 \times ( P(E_0|Z_0)-1) + \mu_1 \left(P(E_1|Z_0)+ P(E_2|Z_0) +P(..|Z_0) -1 \right) \nonumber \\
& &\displaystyle  - \lambda_1 P(E_1|Z_0) \ln(P(E_1|Z_0)) - \lambda_2 P(E_2|Z_0)\ln(P(E_2|Z_0)) - \lambda_{..} P(..|Z_0) \ln (P(..|Z_0)),
\end{eqnarray}
from which we will explain the use in the next part. Due to the update from $Z_0$ to $Z_1$, we should not forget that $P(E_0|Z) \rightarrow P(E_0|Z_0)$ as we wrote previously in Eq.(\ref{ex:fundamental_ignorance}) : the memory of the truth of the proposition is retained during the update. 

\section{Minimising one's ignorance : the Maximum Entropy Principle on Knowledge}

\subsection{Results of the derivatives}
Now let's say Howard is wondering about the coin, and more importantly, how to determine what it is. He is therefore asking how to minimize his Ignorance. To do that, he has different ways : using the functional derivatives, he can derive the Ignorance  
\begin{itemize}
\item \textbf{with respect to $\mu_0$ : } using only the Ignorance of what is known when he was at his state $Z_0$ ,
\begin{equation}
\left.\dfrac{\delta H}{\delta \mu_0}\right|_{Z_1} = 0 \Leftrightarrow    P(E_0|Z_1) -1 = 0  \hskip0.5truecm \Leftrightarrow \hskip0.5truecm  P(E_0|Z_1) = 1
\end{equation}
giving us back the fact that he knows (and remember) for sure that "Things exist". Notice that we have set $Z_0 \rightarrow Z_1$ as we evaluate the proposition "\`a la lumi\`ere" of his new possible update.

\item \textbf{with respect to $\mu_1$} in the same way and for the same reason,   
\begin{equation} \label{eq:E1E2E_at_Z1}
\left.\dfrac{\delta H}{\delta \mu_1}\right|_{Z_1} = 0 \Leftrightarrow  P(E_1|Z_1)+ P(E_2|Z_1) +P(..|Z_1)-1 = 0 \hskip0.5truecm  \Leftrightarrow  \hskip0.5truecm  P(E_1|Z_1) + P(E_2|Z_1) + P(..|Z_1) = 1
\end{equation}
which is what he knows now about the whole current situation would remain true after the update.
\item \textbf{with respect to $\lambda_i$} : using only the Igorance of what remains to be known for each unknow probability, 
\begin{equation}
\left.\dfrac{\delta H}{\delta \lambda_i}\right|_{Z_1}= 0 \Leftrightarrow - P(E_i|Z_1) \ln(P(E_i|Z_1)) = 0
\end{equation}
which leads to what he can speculate about $E_1$ for instance : 
\begin{itemize}
\item $P(E_1 |Z_1) \rightarrow 0 $ : in this case, if the coin does not exist, he will think, and even more, could assert, as such.
\item $\ln (P(E_1 |Z_1)) = 0 \Leftrightarrow P(E_1 |Z_1) = 1 $ : if he comes across the coin, he will have a proof of its existence. 
\end{itemize}
From Eq.(\ref{eq:E1E2E_at_Z1}), does it mean for instance that, due to $P(E_1 |Z_1) = 1$ we can conclude that (A) : $P(E_2 |Z_1) = P(E_{..} |Z_1) = 0$, that is (B) : $E_2$ and $E_{..}$ are false and therefore neither a dice or something else which is not a coin, exist ? Of course, not. (A) is true but does not imply (B): by minimizing his Ignorance, Howard is kind of updating his knowledge "by anticipating" what should be the answers, going therefore from a state $Z_1$ to a state $Z_2$ (still, at another point in spacetime). Indeed, at the moment he will be asserting that the coin exists, it will be about the coin only and he will have no clues concerning the other propositions. That is to say, he can not update simultaneously his knowledge about different objects due to ... causality : when he (or you) first read the paper about the coin and the dice, the information about the coin came first and then the mecanisms in his brain about the update have first recorded the informations about the coin (and where/when he had the information), and at a second time, the informations about the dice. This means also that, being at the previous state $Z_0$, Howard would have set $P(E_1|Z_0) = 1$, leading to $P(E_1 |Z_1) = 1$ at state $Z_1$ too ! 

\newpage

In this example, 
\begin{itemize}
\item his knowledge will be given by 
\begin{equation}
h_2[E_0,E_1,E_2,.., Z_2(Z_1(Z_0,.)),.] = \mu_0 (P(E_0|Z_1)-1) + \mu_1(P(E_1|Z_1)-1) + \mu_2(P(E_2|Z_1)+P(\bar{E}_2|Z_1)-1)
\end{equation}
\item his doubts will be given by 
\begin{equation} 
\bar{h}_2[E_2,\bar{E}_2,Z_1,..] = - \lambda_1 P(E_2|Z_1) \ln P(E_2|Z_1) - \lambda_2 P(\bar{E}_2|Z_1) \ln P(\bar{E}_2|Z_1)
\end{equation}
\item and his state of knowledge $Z_2$ would be represented by 
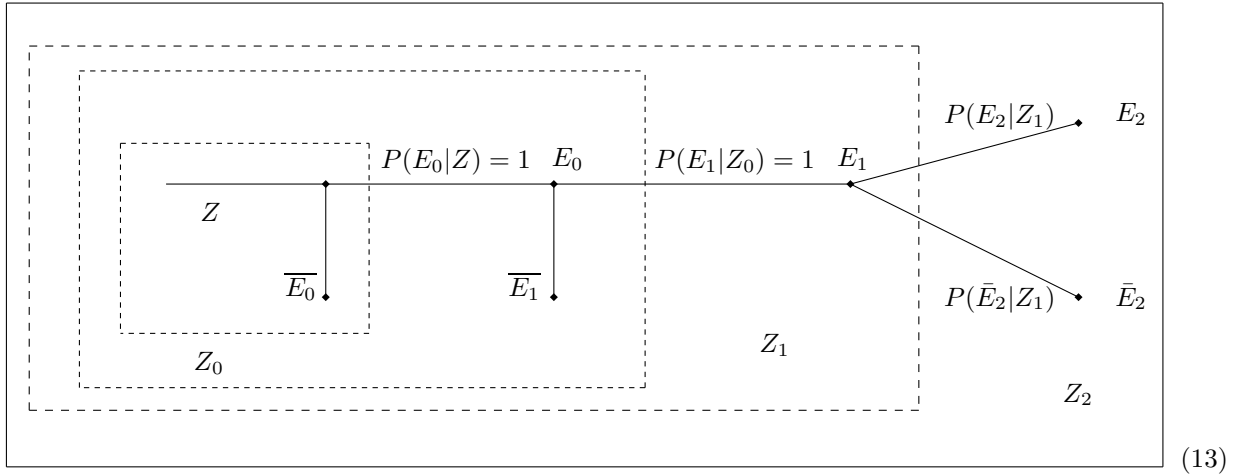
\begin{equation}\label{graph:updateZ2_afterE1}
\begin{tikzpicture}[scale = 3]
\draw (6.2,-0.5)-- (5.2,-0.5);
\draw (5.4,-0.3) node[anchor=north west] {$P(E_0 | Z ) = 1$};
\draw (6.15,-0.3) node[anchor=north west] {$E_0$};
\draw (4.58,-1.2) node[anchor=north west] {$Z_0$};
\draw (5.2,-0.5)-- (5.2,-1);
\draw (4.98,-0.85) node[anchor=north west] {$\overline{E_0}$};
\draw (4.61,-0.54) node[anchor=north west] {$Z$};
\draw [line width=0.4pt,dash pattern=on 2pt off 2pt] (4.3,-0.32)-- (5.39,-0.32);
\draw [line width=0.4pt,dash pattern=on 2pt off 2pt] (5.39,-0.32)-- (5.39,-1.16);
\draw [line width=0.4pt,dash pattern=on 2pt off 2pt] (5.39,-1.16)-- (4.3,-1.16);
\draw [line width=0.4pt,dash pattern=on 2pt off 2pt] (4.3,-0.32)-- (4.3,-1.16);
\draw [dash pattern=on 2pt off 2pt] (4.12,0)-- (6.6,0);
\draw [dash pattern=on 2pt off 2pt] (6.6,-1.4)-- (6.6,0);
\draw [dash pattern=on 2pt off 2pt] (6.6,-1.4)-- (4.12,-1.4);
\draw [dash pattern=on 2pt off 2pt] (4.12,-1.4)-- (4.12,0);
\draw (6.2,-0.5)-- (7.5,-0.5);
\draw (7.4,-0.3) node[anchor=north west] {$E_1$};
\draw (5.2,-0.5)-- (4.5,-0.5);
\draw [dash pattern=on 3pt off 3pt] (3.9,0.11)-- (7.8,0.11);
\draw [dash pattern=on 3pt off 3pt] (7.8,0.11)-- (7.8,-1.5);
\draw [dash pattern=on 3pt off 3pt] (7.8,-1.5)-- (3.9,-1.5);
\draw [dash pattern=on 3pt off 3pt] (3.9,-1.5)-- (3.9,0.11);
\draw (6.2,-0.5)-- (6.2,-1);
\draw (5.96,-0.85) node[anchor=north west] {$\overline{E_1}$};
\draw (6.6,-0.3) node[anchor=north west] {$P(E_1| Z_0 ) = 1$};
\draw (7.06,-1.12) node[anchor=north west] {$Z_1$};
\draw (7.5,-0.5)-- (8.5,-0.23);
\draw (7.5,-0.5)-- (8.5,-1);
\draw (8.62,-0.1) node[anchor=north west] {$E_2$};
\draw (8.62,-0.9) node[anchor=north west] {$\bar{E}_2$};
\draw (3.8,0.3)-- (8.87,0.3);
\draw (8.87,0.3)-- (8.87,-1.75);
\draw (8.87,-1.75)-- (3.8,-1.75);
\draw (3.8,-1.75)-- (3.8,0.3);
\draw (8.39,-1.34) node[anchor=north west] {$Z_2$};
\draw (7.87,-0.1) node[anchor=north west] {$P(E_2| Z_1)$};
\draw (7.87,-0.9) node[anchor=north west] {$P(\bar{E}_2| Z_1 )$};
\begin{scriptsize}
\fill [color=black] (6.2,-0.5) ++(-0.5pt,0 pt) -- ++(0.5pt,0.5pt)--++(0.5pt,-0.5pt)--++(-0.5pt,-0.5pt)--++(-0.5pt,0.5pt);
\fill [color=black] (5.2,-0.5) ++(-0.5pt,0 pt) -- ++(0.5pt,0.5pt)--++(0.5pt,-0.5pt)--++(-0.5pt,-0.5pt)--++(-0.5pt,0.5pt);
\fill [color=black] (5.2,-1) ++(-0.5pt,0 pt) -- ++(0.5pt,0.5pt)--++(0.5pt,-0.5pt)--++(-0.5pt,-0.5pt)--++(-0.5pt,0.5pt);
\fill [color=black] (7.5,-0.5) ++(-0.5pt,0 pt) -- ++(0.5pt,0.5pt)--++(0.5pt,-0.5pt)--++(-0.5pt,-0.5pt)--++(-0.5pt,0.5pt);
\fill [color=black] (6.2,-1) ++(-0.5pt,0 pt) -- ++(0.5pt,0.5pt)--++(0.5pt,-0.5pt)--++(-0.5pt,-0.5pt)--++(-0.5pt,0.5pt);
\fill [color=black] (8.5,-0.23) ++(-0.5pt,0 pt) -- ++(0.5pt,0.5pt)--++(0.5pt,-0.5pt)--++(-0.5pt,-0.5pt)--++(-0.5pt,0.5pt);
\fill [color=black] (8.5,-1) ++(-0.5pt,0 pt) -- ++(0.5pt,0.5pt)--++(0.5pt,-0.5pt)--++(-0.5pt,-0.5pt)--++(-0.5pt,0.5pt);
\end{scriptsize}
\end{tikzpicture}
\end{equation}
\end{itemize}    

\item \textbf{derivating with respect to $P(E_i |Z_n)$} : by interacting with both kind of Ignorances, and leading to a much more general framework if more constraints are added,
\begin{itemize}
\item for $P(E_0,Z)$ : $\left.\dfrac{\delta H}{\delta P(E_0,Z)}\right|_{Z_1}= 0$ naturally due to the constraint setting $P(E_0,Z) =1$ since H does not depend really on $P(E_0|Z)$, and as the derivative of 0 is 0 because knowing that $E_0$ is true leads to no ignorance from the beginning. Otherwise it would mean that $\mu_0 = 0$, that is, Howard has forgotten about $E_0$.
\item for $P(E_i|Z_0)$ : setting $\lambda_i \neq 0$ (otherwise there would be no "unknown" variables) 
\begin{eqnarray}
\left.\dfrac{\delta H}{\delta P(E_i |Z_0)}\right|_{Z_1} = 0 &\Leftrightarrow &  \mu_1 - \lambda_i \left(ln(P(E_i|Z_1) + 1\right) = 0 \\
&\Leftrightarrow & P(E_i|Z_1) = exp{\left(\dfrac{\mu_1}{\lambda_i}-1\right)} \label{eq:deltaHdeltaPEZ1}
\end{eqnarray}
as in \cite{GiffinCaticha} and \cite{Sakth} where a density function $J(x)$, $\displaystyle \int J(x)p(x)dx = C$ as the constraint, was used instead of our discret case $\displaystyle \sum_i p_i = 1$.
\end{itemize}
\end{itemize}

It seems clear enough to see the consequences of derivatives other than those with respect to probabilities, and so we will look at the latter in a simple case where the ignorances have the same weight, that is, setting $\lambda_i = \mu_1$.

\subsection{Case where $\lambda_i = \mu_1$}
In this case, an \textit{a priori} particular one, from 
\begin{eqnarray}
&& H[E_0, E_1,E_2,..,\mu |Z_1(Z_0,x^\mu)] =  \mu_0 (P(E_0|Z_0) -1) \label{eq:H_muequallambda_1}\\
& & +   \mu_1 \left[ \dfrac{{}^{} }{ {}^{} } P(E_1|Z_0) + P(E_2|Z_0) + P(..|Z_0)-1 -  P(E_1|Z_0) \ln(P(E_1|Z_0)) -P(E_2|Z_0)\ln(P(E_2|Z_0)) - P(..|Z_0) \ln (P(..|Z_0))  \right] \nonumber 
\end{eqnarray}
and from Eq.(\ref{eq:deltaHdeltaPEZ1}), if Howard wants to minimise his Ignorance, then 
\begin{equation}
P(E_i|Z_1) = exp{\left(\dfrac{\mu_1}{\lambda_i}-1\right)}  = e^{1-1} = e^0 = 1 
\end{equation}
that is, to assert the existence of the coin or the dice, he should .. find them. Of course, since the beginning this is obvious, but here we show another way to derive it : from $\left.\dfrac{\delta H[E_0,E_1,E_2,..,\lambda|Z_1] }{\delta \mu_1}\right|_{Z_1} = 0 $,
\begin{equation}\label{eq:Z1state_constraint_applied}
P(E_1|Z_1) + P(E_2|Z_1) + P(..|Z_1) - 1  -  P(E_1|Z_1) \ln(P(E_1|Z_1))-P(E_2|Z_1)\ln(P(E_2|Z_1)) - P(..|Z_1) \ln (P(..|Z_1))  = 0
\end{equation}
and after evaluating it with the help of the constraint $P(E_1|Z_1) + P(E_2|Z_1) + P(..|Z_1) - 1 $ (this constraint was considered as true at the previous state, it should also be the case after the update, when one proposition \textbf{is confirmed}, otherwise the whole framework would collapse due to inconsitency),  
\begin{equation} 
 -  P(E_1|Z_1) \ln(P(E_1|Z_1))-P(E_2|Z_1)\ln(P(E_2|Z_1)) - P(..|Z_1) \ln (P(..|Z_1))  = 0 
\end{equation}
which gives, as \textit{a priori} all events are independent, the following result : 
\begin{equation}\label{eq:Z1state_constraint_applied_bis}
P(E_i|Z_1) \rightarrow 0 \,\,\, or \,\,\, P(E_i|Z_1) = 1 \,\,\, for\,\,\ i \in \{1,2,..\}
\end{equation}
not forgetting that, if $P(E_1|Z_1) = 1$, then $P(E_2|Z_1) = 0$ and $P(E_..|Z_1) = 0$, so only the event $E_1$ has been assert as true for Howard, and we recover the case of the Graph.(\ref{graph:updateZ2_afterE1}).

\subsection{Comment on the constraints ($n = 1$) and if they are missing (\textit{i.e.} no knowledge)}

\subsubsection{On the constraints : why $n=1$}

As set in \cite{MyIgnorance}, in Eq.(\ref{eq:ignorance_nulle_EO}) and Eq.(\ref{eq:constraint2}), we use a constraint of the form $0 = \left(1-\displaystyle \sum_i p_i \right)^n$ with $n=1$. However, even it is not yet clear about the value of $n$, using $n>1$ would bring nothing, as, when evaluating the terms with the help of the constraints, because they are constraints, the terms derived from it will be null : 
\begin{equation}
\left(\left(1-\displaystyle \sum_i p_i \right)^n\right)' = n \times p_i \times \left(1-\displaystyle \sum_i p_i \right)^{n-1}\,\,\, \,\,\, \xrightarrow[]{1-\displaystyle \sum_i p_i = 0}\,\,\, \,\,\, 0
\end{equation} 

That is why, in this article, we will set $n =1$ and see what we can know from it.

\subsubsection{If the constraints are missing : case of no Knowledge}
For instance, let us consider in a simple case that $H[p_i,\lambda_i] = - \displaystyle \sum_i \,\, \lambda_i\,\, p_i \times \ln p_i $ with $p_i>0$ or $p_i \rightarrow 0$. Then, from the derivative w.r.t $p_i$ for $i$ in ${1,2, ...} $
\begin{equation} \label{eq:case_of_no_knowledge_consequences}
\left.\dfrac{\delta H}{\delta p_i}\right|_{Z} = 0 \hfill \Leftrightarrow  \hfill -  \lambda_i (\ln p_i+1) = 0 \hfill \Leftrightarrow \hfill \lambda_i = 0 \,\, or \,\, p_i = e^{-1}\hfill {}^{}
\end{equation}
\begin{itemize}
\item if $\lambda_i = 0$, this could mean that, having no knowledge about the event of $p_i$, we do not have an Ignorance about it, whatever the probabilities $p_i$. That is to say, \textit{"I do not have concerns about things I do not know they exist".}\\ \noindent \textit{Je ne m'inqui\`ete pas des choses dont je ne connais pas l'existence.}
\item if $p_i = e^{-1}$, then 
\begin{equation}
H[.] = - \displaystyle \sum_i \,\, \lambda_i\,\, p_i \times \ln p_i  = - \sum_i \,\, \lambda_i\,\, e^{-1} \times \ln(e^{-1}) \hfill = \hfill  e^{-1} \sum_i \lambda_i\hfill {}^{}
\end{equation}
which could mean, related also to the case where maybe $\lambda_i=0$ for some $i$, that we would have there a "fundamental" value for our Ignorance, that is to say \textit{"I know that I am Ignorant but I do not know about what".}\\ \noindent \textit{Je sais que je suis ignorant mais je ne sais pas \`a propos de quoi, autrement dit, j'ignore ce que je ne sais pas.}
\end{itemize}

Both mathematical consequences seem to have philosophical meaning and are in fact used by everyone, at least most of the time.

\section{Anticipation as superposition of configurations : the Bayes formula ? } 

Going back to Howard before he asserts that either $E_1$, $E_2$ or $E_M = \overline{E_1\cup E_2}$ are true, starting from the Graph.(\ref{graph:stateatZ1}), he knows that $E_1$, $E_2$ and $E_M$ are possible. He could therefore wonder in which order he will assert or not their truth and draw the graph as the one in Fig.(\ref{graph:anticipation_stateE1capE2}). In the following, we will look at states where $E_1$ and $E_2$ should be true, in order to see how Bayes formula takes place in this framework.\sk 

However, the following is not considered as clear as we would want, the global view showing quite a complexity about what we should have. As such, we just sketch a possible proof about how we can recover Bayes formula in the simple case of the state $Z_2$, and comment the more complexe cases of $Z_n$, $n>2$. This work is therefore considered to be in progress.

\subsection{The global view}

\begin{figure}[htb]
\definecolor{qqttcc}{rgb}{0.05, 0.11, 0.51}
\begin{tikzpicture}[scale=2]
\draw (2.52,-1.44)-- (1.02,-1.44);
\draw (1.33,-1.08) node[anchor=north west] {$P(E_0 | Z ) = 1$};
\draw (2.34,-1.45) node[anchor=north west] {$E_0$};
\draw (0.18,-2.07) node[anchor=north west] {$Z_0$};
\draw (1.02,-1.44)-- (1.02,-2.24);
\draw (0.7,-2.01) node[anchor=north west] {$\overline{E_0}$};
\draw (0.21,-1.54) node[anchor=north west] {$Z$};
\draw [line width=0.4pt,dash pattern=on 1pt off 1pt] (-0.38,-1.04)-- (0.42,-1.04);
\draw [line width=0.4pt,dash pattern=on 1pt off 1pt] (0.42,-1.04)-- (0.42,-1.84);
\draw [line width=0.4pt,dash pattern=on 1pt off 1pt] (0.42,-1.84)-- (-0.38,-1.84);
\draw [line width=0.4pt,dash pattern=on 1pt off 1pt] (-0.38,-1.04)-- (-0.38,-1.84);
\draw [dash pattern=on 1pt off 1pt] (-0.58,-0.84)-- (2.82,-0.84);
\draw [dash pattern=on 1pt off 1pt] (2.82,-2.44)-- (2.82,-0.84);
\draw [dash pattern=on 1pt off 1pt] (2.82,-2.44)-- (-0.58,-2.44);
\draw [dash pattern=on 1pt off 1pt] (-0.58,-2.44)-- (-0.58,-0.84);
\draw [line width=1.2pt,color=qqttcc] (2.52,-1.44)-- (4.1,-0.43);
\draw [line width=1.2pt,color=qqttcc] (2.52,-1.44)-- (4.1,-1.44);
\draw [line width=1.2pt,color=qqttcc] (2.52,-1.44)-- (4.1,-2.43);
\draw (3.82,-0.1) node[anchor=north west] {$E_1$};
\draw (3.92,-1.08) node[anchor=north west] {$E_2$};
\draw (3.84,-2.47) node[anchor=north west] {$E_{M}$};
\draw (3.15,-0.46) node[anchor=north west] {$P(E_1 | Z_0 )$};
\draw (3.24,-1.12) node[anchor=north west] {$P(E_2 | Z_0)$};
\draw (3.39,-1.69) node[anchor=north west] {$P(E_M | Z_0)$};
\draw (1.02,-1.44)-- (0.1,-1.43);
\draw [dash pattern=on 2pt off 2pt] (-0.9,0.07)-- (4.6,0.07);
\draw [dash pattern=on 2pt off 2pt] (4.6,0.07)-- (4.6,-2.93);
\draw [dash pattern=on 2pt off 2pt] (4.6,-2.93)-- (-0.9,-2.93);
\draw [dash pattern=on 2pt off 2pt] (-0.9,-2.93)-- (-0.9,0.07);
\draw (0.14,-2.54) node[anchor=north west] {$Z_1$};
\draw [line width=1.2pt,color=qqttcc] (4.1,-0.43)-- (5.12,0.67);
\draw [line width=1.2pt,color=qqttcc] (4.1,-0.43)-- (5.12,-0.33);
\draw (5.12,0.67)-- (6.12,0.67);
\draw [line width=1.2pt,color=qqttcc] (5.12,-0.33)-- (6.12,-0.33);
\draw (5.12,0.67)-- (6.12,1.17);
\draw [line width=1.2pt,color=qqttcc] (5.12,-0.33)-- (6.12,0.17);
\draw [line width=1.2pt,color=qqttcc] (4.1,-1.44)-- (5.12,-0.83);
\draw [line width=1.2pt,color=qqttcc] (4.1,-1.44)-- (5.12,-1.83);
\draw [line width=1.2pt,color=qqttcc] (4.1,-2.43)-- (5.12,-2.83);
\draw [line width=1.2pt,color=qqttcc] (4.1,-2.43)-- (5.16,-3.52);
\draw [dash pattern=on 1pt off 1pt] (5.12,-0.83)-- (6.12,-0.83);
\draw [dash pattern=on 1pt off 1pt] (5.12,-0.83)-- (6.12,-1.33);
\draw [line width=1.2pt,color=qqttcc] (5.12,-1.83)-- (6.12,-1.83);
\draw [line width=1.2pt,color=qqttcc] (5.12,-1.83)-- (6.12,-2.33);
\draw [line width=1.2pt,color=qqttcc] (5.12,-2.83)-- (6.12,-2.83);
\draw [line width=1.2pt,color=qqttcc] (5.12,-2.83)-- (6.12,-3.33);
\draw [line width=1.2pt,color=qqttcc] (5.16,-3.52)-- (6.12,-3.83);
\draw [line width=1.2pt,color=qqttcc] (5.16,-3.52)-- (6.12,-4.33);
\draw (5.62,1.17)-- (-1.38,1.17);
\draw (-1.38,1.17)-- (-1.38,-4.33);
\draw (5.62,-4.33)-- (5.62,1.17);
\draw (5.62,-4.33)-- (-1.38,-4.33);
\draw (-1.66,1.39)-- (6.64,1.36);
\draw (6.62,-4.83)-- (-1.69,-4.84);
\draw (-1.66,1.39)-- (-1.69,-4.84);
\draw [dash pattern=on 1pt off 1pt] (6.62,-4.83)-- (6.64,1.36);
\draw (0.1,-3.69) node[anchor=north west] {$Z_2$};
\draw (0.1,-4.38) node[anchor=north west] {$Z_3$};
\draw (0.1,-5) node[anchor=north west] {$Z_4..$};
\draw (4.02,0.52) node[anchor=north west] {$P(E_2 | E_1, Z_1)$};
\draw (4.31,-0.37) node[anchor=north west] {$P(E_M | E_1, Z_1)$};
\draw (4.94,1.01) node[anchor=north west] {$E_2$};
\draw (5.96,0.58) node[anchor=north west] {$E_2$};
\draw (5.15,-3.18) node[anchor=north west] {$E_2$};
\draw (5.05,-2.39) node[anchor=north west] {$E_1$};
\draw (5.81,1.36) node[anchor=north west] {$E_{..}$};
\draw (6.04,-2.49) node[anchor=north west] {$E_2$};
\draw (5,0.06) node[anchor=north west] {$E_{M}$};
\draw (6.01,1.02) node[anchor=north west] {$E_{..}$};
\draw (6.05,0.02) node[anchor=north west] {$E_{..}$};
\draw (4.98,-0.84) node[anchor=north west] {$E_1$};
\draw (6.05,-1.49) node[anchor=north west] {$E_1$};
\draw (6.11,-3.49) node[anchor=north west] {$E_1$};
\draw [line width=1.2pt,color=qqttcc] (6.12,-0.33)-- (7.15,-0.08);
\draw [line width=1.2pt,color=qqttcc] (6.12,-0.33)-- (7.11,-0.54);
\draw (7.28,0.08) node[anchor=north west] {$E_2$};
\draw (6.01,-0.48) node[anchor=north west] {$E_{..}$};
\draw (6.03,-0.98) node[anchor=north west] {$E_{..}$};
\draw (5.01,-1.49) node[anchor=north west] {$E_{M}$};
\draw (6.06,-2.01) node[anchor=north west] {$E_{..}$};
\draw [line width=1.2pt,color=qqttcc] (6.12,-2.33)-- (7.14,-2.16);
\draw [line width=1.2pt,color=qqttcc] (6.12,-2.33)-- (7.12,-2.33);
\draw (7.28,-1.9) node[anchor=north west] {$E_1$};
\draw (6.05,-3.01) node[anchor=north west] {$E_{..}$};
\draw [line width=1.2pt,color=qqttcc] (6.12,-3.33)-- (7.09,-3.15);
\draw [line width=1.2pt,color=qqttcc] (6.12,-3.33)-- (7.12,-3.54);
\draw (7.25,-2.95) node[anchor=north west] {$E_2$};
\draw [line width=1.2pt,color=qqttcc] (7.13,-4.12)-- (6.12,-4.33);
\draw [line width=1.2pt,color=qqttcc] (6.12,-4.33)-- (7.15,-4.53);
\draw (7.29,-3.9) node[anchor=north west] {$E_1$};
\draw [dash pattern=on 1pt off 1pt] (7.15,0.99)-- (6.12,0.67);
\draw [dash pattern=on 1pt off 1pt] (7.1,0.59)-- (6.12,0.67);
\draw [dash pattern=on 1pt off 1pt] (6.12,1.17)-- (7.12,1.53);
\draw [dash pattern=on 1pt off 1pt] (6.12,1.17)-- (7.1,1.23);
\draw [dash pattern=on 1pt off 1pt] (7.14,0.35)-- (6.12,0.17);
\draw [dash pattern=on 1pt off 1pt] (6.12,0.17)-- (7.12,0.17);
\draw [dash pattern=on 1pt off 1pt] (6.12,-0.83)-- (7.12,-0.83);
\draw [dash pattern=on 1pt off 1pt] (6.12,-0.83)-- (7.08,-1.04);
\draw [dash pattern=on 1pt off 1pt] (6.12,-1.33)-- (7.13,-1.21);
\draw [dash pattern=on 1pt off 1pt] (7.13,-1.42)-- (6.12,-1.33);
\draw [dash pattern=on 1pt off 1pt] (6.12,-1.83)-- (7.11,-1.62);
\draw [dash pattern=on 1pt off 1pt] (6.12,-1.83)-- (7.12,-1.83);
\draw [dash pattern=on 1pt off 1pt] (6.12,-2.83)-- (7.09,-2.66);
\draw [dash pattern=on 1pt off 1pt] (6.12,-2.83)-- (7.12,-2.83);
\draw [dash pattern=on 1pt off 1pt] (6.12,-3.83)-- (7.12,-3.71);
\draw [dash pattern=on 1pt off 1pt] (6.12,-3.83)-- (7.12,-3.9);
\draw (5.98,-4.31) node[anchor=north west] {$E_{..}$};
\draw (7.19,-0.34) node[anchor=north west] {$.. E_2 ..$};
\draw (7.24,-2.16) node[anchor=north west] {$.. E_1 ..$};
\draw (7.21,-3.35) node[anchor=north west] {$.. E_2 ..$};
\draw (7.27,-4.34) node[anchor=north west] {$.. E_1 ..$};
\draw [color=qqttcc] (4.1,-2.43)-- (4.53,-3.86);
\draw [color=qqttcc] (4.53,-3.86)-- (5.71,-4.48);
\draw [color=qqttcc] (4.53,-3.86)-- (5.18,-4.51);
\draw [color=qqttcc] (4.53,-3.86)-- (4.2,-4.54);
\draw (5.68,-4.58) node[anchor=north west] {$E_1..E_2..$};
\draw (4.2,-3.74) node[anchor=north west] {$E_{..}$};
\draw (5,-4.6) node[anchor=north west] {$E_2..E_1..$};
\draw (3.7,-4.52) node[anchor=north west] {$..E_1/E_2..E_2/E_1..$};
\begin{scriptsize}
\fill [color=black] (2.52,-1.44) ++(-1.5pt,0 pt) -- ++(1.5pt,1.5pt)--++(1.5pt,-1.5pt)--++(-1.5pt,-1.5pt)--++(-1.5pt,1.5pt);
\fill [color=black] (1.02,-1.44) ++(-1.5pt,0 pt) -- ++(1.5pt,1.5pt)--++(1.5pt,-1.5pt)--++(-1.5pt,-1.5pt)--++(-1.5pt,1.5pt);
\fill [color=black] (1.02,-2.24) ++(-1.5pt,0 pt) -- ++(1.5pt,1.5pt)--++(1.5pt,-1.5pt)--++(-1.5pt,-1.5pt)--++(-1.5pt,1.5pt);
\fill [color=black] (4.1,-0.43) ++(-1.5pt,0 pt) -- ++(1.5pt,1.5pt)--++(1.5pt,-1.5pt)--++(-1.5pt,-1.5pt)--++(-1.5pt,1.5pt);
\fill [color=black] (4.1,-1.44) ++(-1.5pt,0 pt) -- ++(1.5pt,1.5pt)--++(1.5pt,-1.5pt)--++(-1.5pt,-1.5pt)--++(-1.5pt,1.5pt);
\fill [color=black] (4.1,-2.43) ++(-1.5pt,0 pt) -- ++(1.5pt,1.5pt)--++(1.5pt,-1.5pt)--++(-1.5pt,-1.5pt)--++(-1.5pt,1.5pt);
\fill [color=black] (5.12,0.67) ++(-1.5pt,0 pt) -- ++(1.5pt,1.5pt)--++(1.5pt,-1.5pt)--++(-1.5pt,-1.5pt)--++(-1.5pt,1.5pt);
\fill [color=black] (5.12,-0.33) ++(-1.5pt,0 pt) -- ++(1.5pt,1.5pt)--++(1.5pt,-1.5pt)--++(-1.5pt,-1.5pt)--++(-1.5pt,1.5pt);
\fill [color=black] (6.12,0.67) ++(-1.5pt,0 pt) -- ++(1.5pt,1.5pt)--++(1.5pt,-1.5pt)--++(-1.5pt,-1.5pt)--++(-1.5pt,1.5pt);
\fill [color=black] (6.12,-0.33) ++(-1.5pt,0 pt) -- ++(1.5pt,1.5pt)--++(1.5pt,-1.5pt)--++(-1.5pt,-1.5pt)--++(-1.5pt,1.5pt);
\fill [color=black] (6.12,1.17) ++(-1.5pt,0 pt) -- ++(1.5pt,1.5pt)--++(1.5pt,-1.5pt)--++(-1.5pt,-1.5pt)--++(-1.5pt,1.5pt);
\fill [color=black] (6.12,0.17) ++(-1.5pt,0 pt) -- ++(1.5pt,1.5pt)--++(1.5pt,-1.5pt)--++(-1.5pt,-1.5pt)--++(-1.5pt,1.5pt);
\fill [color=black] (5.12,-0.83) ++(-1.5pt,0 pt) -- ++(1.5pt,1.5pt)--++(1.5pt,-1.5pt)--++(-1.5pt,-1.5pt)--++(-1.5pt,1.5pt);
\fill [color=black] (5.12,-1.83) ++(-1.5pt,0 pt) -- ++(1.5pt,1.5pt)--++(1.5pt,-1.5pt)--++(-1.5pt,-1.5pt)--++(-1.5pt,1.5pt);
\fill [color=black] (5.12,-2.83) ++(-1.5pt,0 pt) -- ++(1.5pt,1.5pt)--++(1.5pt,-1.5pt)--++(-1.5pt,-1.5pt)--++(-1.5pt,1.5pt);
\fill [color=black] (5.16,-3.52) ++(-1.5pt,0 pt) -- ++(1.5pt,1.5pt)--++(1.5pt,-1.5pt)--++(-1.5pt,-1.5pt)--++(-1.5pt,1.5pt);
\fill [color=black] (6.12,-0.83) ++(-1.5pt,0 pt) -- ++(1.5pt,1.5pt)--++(1.5pt,-1.5pt)--++(-1.5pt,-1.5pt)--++(-1.5pt,1.5pt);
\fill [color=black] (6.12,-1.33) ++(-1.5pt,0 pt) -- ++(1.5pt,1.5pt)--++(1.5pt,-1.5pt)--++(-1.5pt,-1.5pt)--++(-1.5pt,1.5pt);
\fill [color=black] (6.12,-1.83) ++(-1.5pt,0 pt) -- ++(1.5pt,1.5pt)--++(1.5pt,-1.5pt)--++(-1.5pt,-1.5pt)--++(-1.5pt,1.5pt);
\fill [color=black] (6.12,-2.33) ++(-1.5pt,0 pt) -- ++(1.5pt,1.5pt)--++(1.5pt,-1.5pt)--++(-1.5pt,-1.5pt)--++(-1.5pt,1.5pt);
\fill [color=black] (6.12,-2.83) ++(-1.5pt,0 pt) -- ++(1.5pt,1.5pt)--++(1.5pt,-1.5pt)--++(-1.5pt,-1.5pt)--++(-1.5pt,1.5pt);
\fill [color=black] (6.12,-3.33) ++(-1.5pt,0 pt) -- ++(1.5pt,1.5pt)--++(1.5pt,-1.5pt)--++(-1.5pt,-1.5pt)--++(-1.5pt,1.5pt);
\fill [color=black] (6.12,-3.83) ++(-1.5pt,0 pt) -- ++(1.5pt,1.5pt)--++(1.5pt,-1.5pt)--++(-1.5pt,-1.5pt)--++(-1.5pt,1.5pt);
\fill [color=black] (6.12,-4.33) ++(-1.5pt,0 pt) -- ++(1.5pt,1.5pt)--++(1.5pt,-1.5pt)--++(-1.5pt,-1.5pt)--++(-1.5pt,1.5pt);
\fill [color=black] (7.15,-0.08) ++(-1.5pt,0 pt) -- ++(1.5pt,1.5pt)--++(1.5pt,-1.5pt)--++(-1.5pt,-1.5pt)--++(-1.5pt,1.5pt);
\fill [color=black] (7.11,-0.54) ++(-1.5pt,0 pt) -- ++(1.5pt,1.5pt)--++(1.5pt,-1.5pt)--++(-1.5pt,-1.5pt)--++(-1.5pt,1.5pt);
\fill [color=black] (7.14,-2.16) ++(-1.5pt,0 pt) -- ++(1.5pt,1.5pt)--++(1.5pt,-1.5pt)--++(-1.5pt,-1.5pt)--++(-1.5pt,1.5pt);
\fill [color=black] (7.12,-2.33) ++(-1.5pt,0 pt) -- ++(1.5pt,1.5pt)--++(1.5pt,-1.5pt)--++(-1.5pt,-1.5pt)--++(-1.5pt,1.5pt);
\fill [color=black] (7.09,-3.15) ++(-1.5pt,0 pt) -- ++(1.5pt,1.5pt)--++(1.5pt,-1.5pt)--++(-1.5pt,-1.5pt)--++(-1.5pt,1.5pt);
\fill [color=black] (7.12,-3.54) ++(-1.5pt,0 pt) -- ++(1.5pt,1.5pt)--++(1.5pt,-1.5pt)--++(-1.5pt,-1.5pt)--++(-1.5pt,1.5pt);
\fill [color=black] (7.13,-4.12) ++(-1.5pt,0 pt) -- ++(1.5pt,1.5pt)--++(1.5pt,-1.5pt)--++(-1.5pt,-1.5pt)--++(-1.5pt,1.5pt);
\fill [color=black] (7.15,-4.53) ++(-1.5pt,0 pt) -- ++(1.5pt,1.5pt)--++(1.5pt,-1.5pt)--++(-1.5pt,-1.5pt)--++(-1.5pt,1.5pt);
\fill [color=black] (7.15,0.99) ++(-1.5pt,0 pt) -- ++(1.5pt,1.5pt)--++(1.5pt,-1.5pt)--++(-1.5pt,-1.5pt)--++(-1.5pt,1.5pt);
\fill [color=black] (7.1,0.59) ++(-1.5pt,0 pt) -- ++(1.5pt,1.5pt)--++(1.5pt,-1.5pt)--++(-1.5pt,-1.5pt)--++(-1.5pt,1.5pt);
\fill [color=black] (7.12,1.53) ++(-1.5pt,0 pt) -- ++(1.5pt,1.5pt)--++(1.5pt,-1.5pt)--++(-1.5pt,-1.5pt)--++(-1.5pt,1.5pt);
\fill [color=black] (7.1,1.23) ++(-1.5pt,0 pt) -- ++(1.5pt,1.5pt)--++(1.5pt,-1.5pt)--++(-1.5pt,-1.5pt)--++(-1.5pt,1.5pt);
\fill [color=black] (7.14,0.35) ++(-1.5pt,0 pt) -- ++(1.5pt,1.5pt)--++(1.5pt,-1.5pt)--++(-1.5pt,-1.5pt)--++(-1.5pt,1.5pt);
\fill [color=black] (7.12,0.17) ++(-1.5pt,0 pt) -- ++(1.5pt,1.5pt)--++(1.5pt,-1.5pt)--++(-1.5pt,-1.5pt)--++(-1.5pt,1.5pt);
\fill [color=black] (7.12,-0.83) ++(-1.5pt,0 pt) -- ++(1.5pt,1.5pt)--++(1.5pt,-1.5pt)--++(-1.5pt,-1.5pt)--++(-1.5pt,1.5pt);
\fill [color=black] (7.08,-1.04) ++(-1.5pt,0 pt) -- ++(1.5pt,1.5pt)--++(1.5pt,-1.5pt)--++(-1.5pt,-1.5pt)--++(-1.5pt,1.5pt);
\fill [color=black] (7.13,-1.21) ++(-1.5pt,0 pt) -- ++(1.5pt,1.5pt)--++(1.5pt,-1.5pt)--++(-1.5pt,-1.5pt)--++(-1.5pt,1.5pt);
\fill [color=black] (7.13,-1.42) ++(-1.5pt,0 pt) -- ++(1.5pt,1.5pt)--++(1.5pt,-1.5pt)--++(-1.5pt,-1.5pt)--++(-1.5pt,1.5pt);
\fill [color=black] (7.11,-1.62) ++(-1.5pt,0 pt) -- ++(1.5pt,1.5pt)--++(1.5pt,-1.5pt)--++(-1.5pt,-1.5pt)--++(-1.5pt,1.5pt);
\fill [color=black] (7.12,-1.83) ++(-1.5pt,0 pt) -- ++(1.5pt,1.5pt)--++(1.5pt,-1.5pt)--++(-1.5pt,-1.5pt)--++(-1.5pt,1.5pt);
\fill [color=black] (7.09,-2.66) ++(-1.5pt,0 pt) -- ++(1.5pt,1.5pt)--++(1.5pt,-1.5pt)--++(-1.5pt,-1.5pt)--++(-1.5pt,1.5pt);
\fill [color=black] (7.12,-2.83) ++(-1.5pt,0 pt) -- ++(1.5pt,1.5pt)--++(1.5pt,-1.5pt)--++(-1.5pt,-1.5pt)--++(-1.5pt,1.5pt);
\fill [color=black] (7.12,-3.71) ++(-1.5pt,0 pt) -- ++(1.5pt,1.5pt)--++(1.5pt,-1.5pt)--++(-1.5pt,-1.5pt)--++(-1.5pt,1.5pt);
\fill [color=black] (7.12,-3.9) ++(-1.5pt,0 pt) -- ++(1.5pt,1.5pt)--++(1.5pt,-1.5pt)--++(-1.5pt,-1.5pt)--++(-1.5pt,1.5pt);
\fill [color=black] (4.53,-3.86) ++(-1.5pt,0 pt) -- ++(1.5pt,1.5pt)--++(1.5pt,-1.5pt)--++(-1.5pt,-1.5pt)--++(-1.5pt,1.5pt);
\fill [color=black] (5.71,-4.48) ++(-1.5pt,0 pt) -- ++(1.5pt,1.5pt)--++(1.5pt,-1.5pt)--++(-1.5pt,-1.5pt)--++(-1.5pt,1.5pt);
\fill [color=black] (5.18,-4.51) ++(-1.5pt,0 pt) -- ++(1.5pt,1.5pt)--++(1.5pt,-1.5pt)--++(-1.5pt,-1.5pt)--++(-1.5pt,1.5pt);
\fill [color=black] (4.2,-4.54) ++(-1.5pt,0 pt) -- ++(1.5pt,1.5pt)--++(1.5pt,-1.5pt)--++(-1.5pt,-1.5pt)--++(-1.5pt,1.5pt);
\end{scriptsize}
\end{tikzpicture}
\caption{Anticipating what's next : in blue, paths leading to a state where both $E_1$ and $E_2$ are true (at lot are missing) in the general cas (if 4 updates, there are 12 paths containing $E_1$ and $E_2$} 
\label{graph:anticipation_stateE1capE2}
\end{figure}
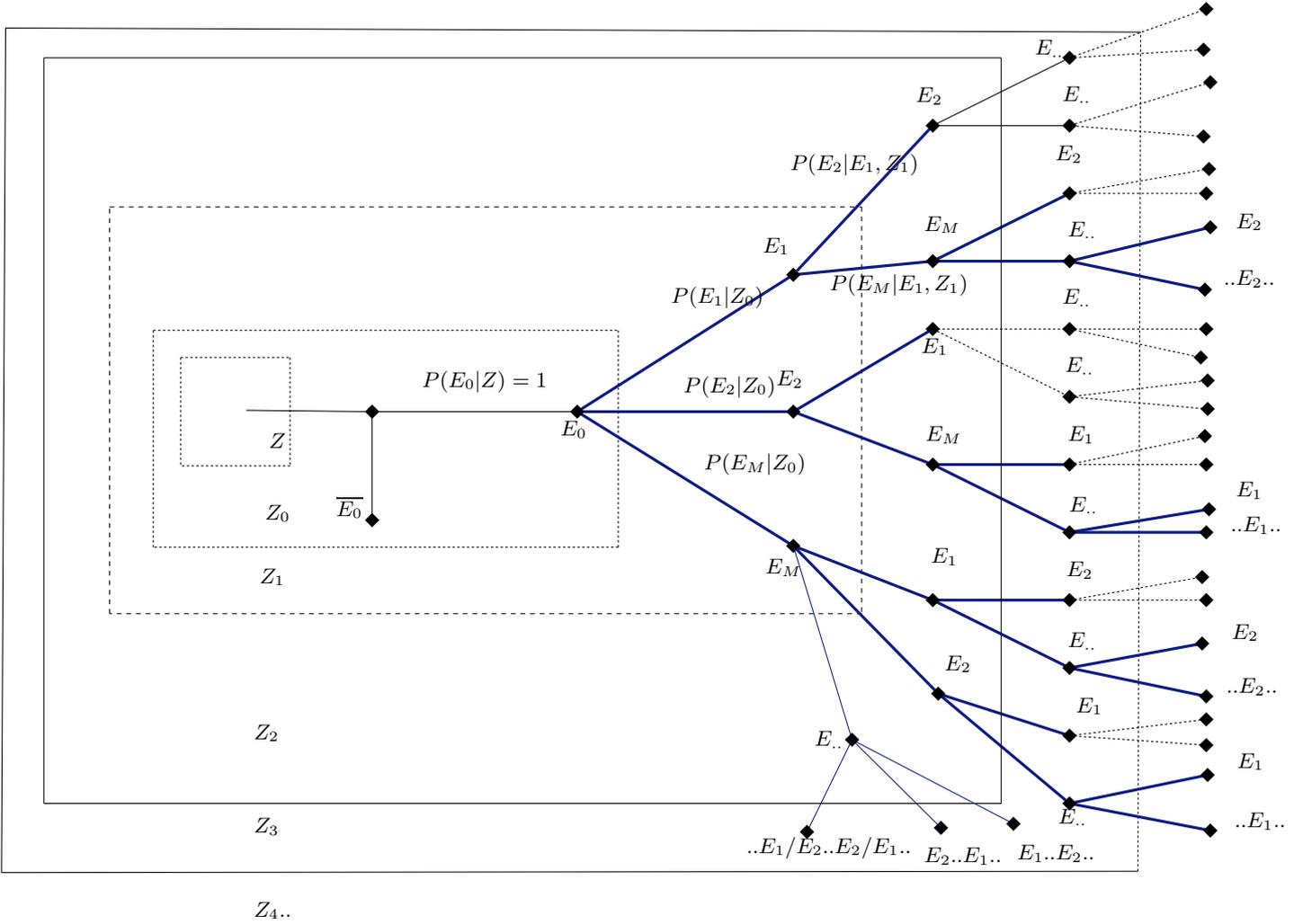

Looking at Fig.(\ref{graph:anticipation_stateE1capE2}), we see that the states which verifies $E_1$ and $E_2$ have either the probabilitis $P(E_1 \cap E_2)$ or $P(E_2 \cap E_1)$ such that 
\begin{equation}
P(E_1 \cap E_2) = \displaystyle \sum_{l,m,n} P\left( \,\, \left( \prod_{i=1}^l E_{..} \cap \right) \,\, E_1 \,\,   \left( \prod_{j=1}^m E_{..} \cap \right) \,\,E_2 \,\, \left( \prod_{k=1}^n \cap  E_{..} \right) \,\, \right) = P(..E_1..E_2..)
\end{equation}
and 
\begin{equation}
P(E_2 \cap E_1) = \displaystyle \sum_{l,m,n} P\left( \,\, \left( \prod_{i=1}^l E_{..} \cap \right) \,\, E_2 \,\,   \left( \prod_{j=1}^m E_{..} \cap \right) \,\,E_1 \,\, \left( \prod_{k=1}^n \cap  E_{..} \right) \,\, \right) = P(..E_2..E_1..)
\end{equation}

\begin{proposition}
The number of paths $u_n$ containing $E_1$ and $E_2$ at the $Z_n$ state, follows the recurrence relation 
\begin{equation}
u_{n+1} = u_n + 2n, \hskip0.5truecm u_1 = 0
\end{equation}
whose solution is 
\begin{equation}
u_n = n^2 - n 
\end{equation}
As $u_n = n(n-1)$ it will always be an even number as expected due to the symmetry $E_1 \leftrightarrow E_2$.
\end{proposition}

\begin{minipage}{0.07\textwidth}
\textit{Proof} 
\end{minipage}\hfill \begin{minipage}{0.92\textwidth}
If we denote $(u_n)$ the sequence caracterising the number of paths having both $E_1$ and $E_2$ at the state $Z_n$, then 
\begin{itemize}
\item at $Z_1$, there are $u_1 = 0$ paths 
\item at $Z_2$, there are $u_2 = 2$ paths (0+2)
\item at $Z_3$, there are $u_3 = 6$ paths (2+4)
\item at $Z_4$, there are $u_4 = 12$ paths (6+6) 
\item at $Z_5$, there are $u_5 = 20$ paths (12+8)
\end{itemize}
from which we can see the following recurrence relation $u_{n+1} = u_n + 2n$ for the number of paths containing both $E_1$ and $E_2$, half of them concern the $E_1,E_2$ order and the other half the order $E_2,E_2$. As the recurrence relation is of order 1 due to the term $2n$, the expression of $u_n$, using the polynomial method, is the second order $u_n = \alpha n^2+\beta n + \gamma$, and a quick resolution with 
\begin{itemize}
\item $u_1 = 0 = \alpha + \beta + \gamma \Leftrightarrow \gamma = -\alpha - \beta$
\item $u_2 = 2 = 4\alpha + 2\beta + \gamma = 3\alpha + \beta  \Leftrightarrow \beta = 2 - 3 \alpha $ and $\gamma = 2\alpha -2$
\item $u_3 = 6 = 9 \alpha + 3(2-3\alpha) +(2\alpha -2) \Leftrightarrow (2\alpha -2)  = 0 \Leftrightarrow \alpha = 1$, and thus $\beta = 2-3 = -1$ and $\gamma = 2-2 = 0$
\end{itemize}

\end{minipage} \sk 

However, due to the complexity it raises, we will focus now on the simple cases $Z_2$ and $Z_3$.

\subsection{Case of 2 possible updates after $E_0$ : $Z_2$ state }

\begin{figure}
\definecolor{qqttcc}{rgb}{0.05, 0.11, 0.51}
\begin{tikzpicture}[scale=1.7]
\draw (8.18,-1.74)-- (6.68,-1.74);
\draw (7,-1.36) node[anchor=north west] {$P(E_0 | Z ) = 1$};
\draw (8,-1.81) node[anchor=north west] {$E_0$};
\draw (5.84,-2.34) node[anchor=north west] {$Z_0$};
\draw (6.68,-1.74)-- (6.68,-2.54);
\draw (6.3,-2.27) node[anchor=north west] {$\overline{E_0}$};
\draw (5.87,-1.82) node[anchor=north west] {$Z$};
\draw [line width=0.4pt,dash pattern=on 1pt off 1pt] (5.28,-1.34)-- (6.08,-1.34);
\draw [line width=0.4pt,dash pattern=on 1pt off 1pt] (6.08,-1.34)-- (6.08,-2.14);
\draw [line width=0.4pt,dash pattern=on 1pt off 1pt] (6.08,-2.14)-- (5.28,-2.14);
\draw [line width=0.4pt,dash pattern=on 1pt off 1pt] (5.28,-1.34)-- (5.28,-2.14);
\draw [dash pattern=on 1pt off 1pt] (5.08,-1.14)-- (8.48,-1.14);
\draw [dash pattern=on 1pt off 1pt] (8.48,-2.74)-- (8.48,-1.14);
\draw [dash pattern=on 1pt off 1pt] (8.48,-2.74)-- (5.08,-2.74);
\draw [dash pattern=on 1pt off 1pt] (5.08,-2.74)-- (5.08,-1.14);
\draw [line width=1.2pt,color=qqttcc] (8.18,-1.74)-- (9.76,-0.73);
\draw [line width=1.2pt,color=qqttcc] (8.18,-1.74)-- (9.76,-1.74);
\draw [line width=1.2pt,dash pattern=on 1pt off 1pt] (8.18,-1.74)-- (9.76,-2.73);
\draw (9.48,-0.38) node[anchor=north west] {$E_1$};
\draw (9.58,-1.36) node[anchor=north west] {$E_2$};
\draw (9.49,-2.73) node[anchor=north west] {$E_{M}$};
\draw (9.7,-2.5) node[anchor=north west] {{$\star$}};
\draw (8.6,-0.73) node[anchor=north west] {$P(E_1 | Z_0 )$};
\draw (8.6,-1.4) node[anchor=north west] {$P(E_2 | Z_0)$};
\draw (9.05,-1.96) node[anchor=north west] {$P(E_M | Z_0)$};
\draw (6.68,-1.74)-- (5.76,-1.73);
\draw [dotted] (4.76,-0.23)-- (10.26,-0.23);
\draw [dotted] (10.26,-0.23)-- (10.26,-3.23);
\draw [dotted] (10.26,-3.23)-- (4.76,-3.23);
\draw [dotted] (4.76,-3.23)-- (4.76,-0.23);
\draw (5.8,-2.81) node[anchor=north west] {$Z_1$};
\draw [line width=1.2pt,color=qqttcc] (9.76,-0.73)-- (10.91,-0.08);
\draw [line width=1.2pt,dash pattern=on 1pt off 1pt] (9.76,-0.73)-- (10.97,-0.73);
\draw [line width=1.2pt,color=qqttcc] (9.76,-1.74)-- (10.81,-1.48);
\draw [line width=1.2pt,dash pattern=on 1pt off 1pt] (9.76,-1.74)-- (10.73,-2.03);
\draw [line width=1.2pt,dash pattern=on 1pt off 1pt] (9.76,-2.73)-- (10.74,-2.63);
\draw [line width=1.2pt,dash pattern=on 1pt off 1pt] (9.76,-2.73)-- (10.74,-3.04);
\draw (5.82,-3.3) node[anchor=north west] {$Z_2$};
\draw (9.68,0.24) node[anchor=north west] {$P(E_2 | E_1, Z_1)$};
\draw (10.22,-0.76) node[anchor=north west] {$P(\overline{E_2} | E_1, Z_1)$};
\draw (11.03,0.11) node[anchor=north west] {$E_2$};
\draw (10.88,-2.82) node[anchor=north west] {$E_2$};
\draw (10.9,-2.38) node[anchor=north west] {$E_1$};
\draw (10.98,-1.25) node[anchor=north west] {$E_1$};
\draw [dash pattern=on 1pt off 1pt] (9.76,-2.73)-- (10.63,-3.5);
\draw (10.81,-3.28) node[anchor=north west] {$E_{..}$};
\draw (11.13,-0.53) node[anchor=north west] {$\overline{E_2}$};
\draw (10.95,-1.77) node[anchor=north west] {$\overline{E_1}$};
\draw (9.84,-1.16) node[anchor=north west] {$P(E_1 | E_2, Z_1)$};
\begin{scriptsize}
\fill [color=black] (8.18,-1.74) ++(-1.5pt,0 pt) -- ++(1.5pt,1.5pt)--++(1.5pt,-1.5pt)--++(-1.5pt,-1.5pt)--++(-1.5pt,1.5pt);
\fill [color=black] (6.68,-1.74) ++(-1.5pt,0 pt) -- ++(1.5pt,1.5pt)--++(1.5pt,-1.5pt)--++(-1.5pt,-1.5pt)--++(-1.5pt,1.5pt);
\fill [color=black] (6.68,-2.54) ++(-1.5pt,0 pt) -- ++(1.5pt,1.5pt)--++(1.5pt,-1.5pt)--++(-1.5pt,-1.5pt)--++(-1.5pt,1.5pt);
\fill [color=black] (9.76,-0.73) ++(-1.5pt,0 pt) -- ++(1.5pt,1.5pt)--++(1.5pt,-1.5pt)--++(-1.5pt,-1.5pt)--++(-1.5pt,1.5pt);
\fill [color=black] (9.76,-1.74) ++(-1.5pt,0 pt) -- ++(1.5pt,1.5pt)--++(1.5pt,-1.5pt)--++(-1.5pt,-1.5pt)--++(-1.5pt,1.5pt);
\fill [color=black] (9.76,-2.73) ++(-1.5pt,0 pt) -- ++(1.5pt,1.5pt)--++(1.5pt,-1.5pt)--++(-1.5pt,-1.5pt)--++(-1.5pt,1.5pt);
\fill [color=black] (10.91,-0.08) ++(-1.5pt,0 pt) -- ++(1.5pt,1.5pt)--++(1.5pt,-1.5pt)--++(-1.5pt,-1.5pt)--++(-1.5pt,1.5pt);
\fill [color=black] (10.97,-0.73) ++(-1.5pt,0 pt) -- ++(1.5pt,1.5pt)--++(1.5pt,-1.5pt)--++(-1.5pt,-1.5pt)--++(-1.5pt,1.5pt);
\fill [color=black] (10.81,-1.48) ++(-1.5pt,0 pt) -- ++(1.5pt,1.5pt)--++(1.5pt,-1.5pt)--++(-1.5pt,-1.5pt)--++(-1.5pt,1.5pt);
\fill [color=black] (10.73,-2.03) ++(-1.5pt,0 pt) -- ++(1.5pt,1.5pt)--++(1.5pt,-1.5pt)--++(-1.5pt,-1.5pt)--++(-1.5pt,1.5pt);
\fill [color=black] (10.74,-2.63) ++(-1.5pt,0 pt) -- ++(1.5pt,1.5pt)--++(1.5pt,-1.5pt)--++(-1.5pt,-1.5pt)--++(-1.5pt,1.5pt);
\fill [color=black] (10.74,-3.04) ++(-1.5pt,0 pt) -- ++(1.5pt,1.5pt)--++(1.5pt,-1.5pt)--++(-1.5pt,-1.5pt)--++(-1.5pt,1.5pt);
\fill [color=black] (10.63,-3.5) ++(-1.5pt,0 pt) -- ++(1.5pt,1.5pt)--++(1.5pt,-1.5pt)--++(-1.5pt,-1.5pt)--++(-1.5pt,1.5pt);
\end{scriptsize}
\end{tikzpicture}
\caption{Case of 2 possible updates after $E_0$ : 2 paths contain $E_1$ and $E_2$} 
\label{fig:2updatesafterE0}
\end{figure}
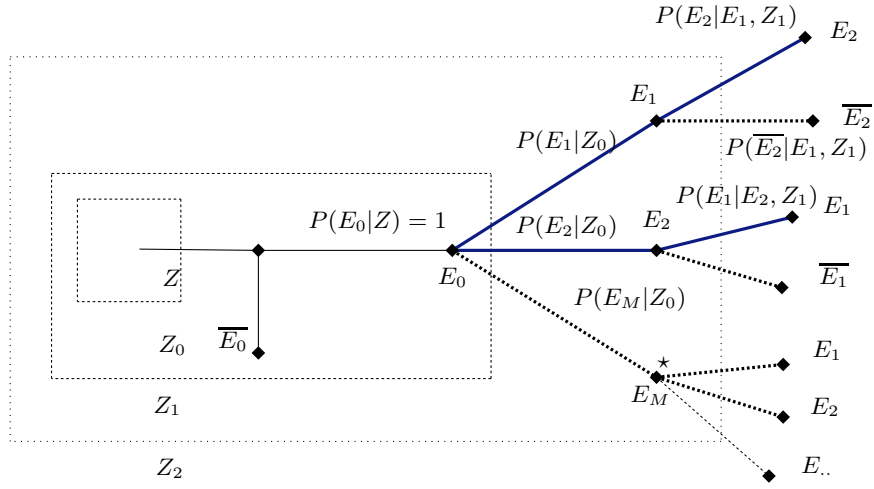

In Fig.(\ref{fig:2updatesafterE0}), we would know that, for the constraints of interest

\begin{itemize}
\item $\gamma \times (P(E_1|Z_0) + P(E_2|Z_0) + P(E_M|Z_0) - 1 ) = 0 $
\item $\alpha P(E_1|Z_0)\times (P(E_2|E_1, Z_1) +  P(E_{..} |E_1, Z_1) - 1) = 0$
\item $\beta P(E_2|Z_0)\times (P(E_1|E_2, Z_1) +  P(E_{..} |E_2, Z_1) - 1) = 0$
\end{itemize} 

but also that 
\begin{equation}\label{eq:IgnoranceprevisionZ2Constraint1}
P(E_1,E_2|Z_1) = k_1 P(E_1 \cap E_2|Z_1) + k_2 P(E_2 \cap E_1|Z_1) = C,
\end{equation}
where $C$ is a constant, and we would \textit{a priori} think that $\displaystyle \sum_i k_i = 1 + 1 = u_2 = 2$. Therefore, we think that Ignorance should have an expression as 
\begin{eqnarray}
&& H[. | Z_2] =  \gamma \times (P(E_1|Z_0) + P(E_2|Z_0) + P(E_M|Z_0) - 1 ) \label{eq:IgnoranceprevisionZ20}\\
&& + \alpha P(E_1|Z_0)\times (P(E_2|E_1, Z_1) +  P(E_{..} |E_1, Z_1) - 1) + \beta P(E_2|Z_0)\times (P(E_1|E_2, Z_1) +  P(E_{..} |E_2, Z_1) - 1)  + ... \label{eq:IgnoranceprevisionZ2A} \\
& & - \lambda_1 P(E_1|Z_0) \ln P(E_1|Z_0) - \lambda_2 P(E_2|Z_0) \ln P(E_2|Z_0) - .... \label{eq:IgnoranceprevisionZ2B}  \\
&& + \delta (k_1 P(E_1 \cap E_2|Z_1) + k_2 P(E_2 \cap E_1|Z_1) -C )\label{eq:IgnoranceprevisionZ2C} \\
&& - \mu_1 P(E_1 \cap E_2|Z_1) \ln P(E_1 \cap E_2|Z_1) - \mu_2 P(E_2 \cap E_1|Z_1) \ln P(E_2 \cap E_1|Z_1)\label{eq:IgnoranceprevisionZ2D}
\end{eqnarray} 
where Eq.(\ref{eq:IgnoranceprevisionZ20}) and Eq.(\ref{eq:IgnoranceprevisionZ2A}) deal with constraints on each branch of the tree, and Eq.(\ref{eq:IgnoranceprevisionZ2B}) with their associated Ignorance, and Eq.(\ref{eq:IgnoranceprevisionZ2C}) and Eq.(\ref{eq:IgnoranceprevisionZ2D}) do the same but for $P(E_1,E_2)$, and it should be possible to express them with terms in Eq.(\ref{eq:IgnoranceprevisionZ20}), Eq.(\ref{eq:IgnoranceprevisionZ2A}) and Eq.(\ref{eq:IgnoranceprevisionZ2B}). 

Therefore 
\begin{eqnarray}
\dfrac{\delta H[.]}{\delta P(E_1\cap E_2)} = 0 &\Leftrightarrow&  \delta k_1 - \mu_1 (\ln (P(E_1\cap E_2))-1 ) = 0 \Leftrightarrow \delta k_1 = \mu_1 (\ln (P(E_1\cap E_2))-1 ) \\
\dfrac{\delta H[.]}{\delta P(E_2\cap E_1)} = 0 &\Leftrightarrow&  \delta k_2 - \mu_1 (\ln (P(E_2\cap E_1))-1 ) = 0 \Leftrightarrow \delta k_2 = \mu_2 (\ln (P(E_2\cap E_1))-1 )
\end{eqnarray} 

Setting $\mu_1 = \mu_2 = \mu $ as both $P(E_1\cap E_2)$ and $P(E_2\cap E_1)$ are linked with the constraint in Eq.(\ref{eq:IgnoranceprevisionZ2Constraint1}), but also that $k_1 = k_2 = 1$ as each probability is given by 1 branch, we have 
\begin{eqnarray}
&& \delta = \mu (\ln (P(E_1\cap E_2))-1 ) = \mu (\ln (P(E_2\cap E_1))-1 ) \\ 
&\Leftrightarrow & P(E_1\cap E_2) = P(E_2\cap E_1) \\ 
&\Leftrightarrow & P(E_1|Z_0) \times P(E_2|E_1,Z_1) = P(E_2|Z_0) \times P(E_1|E_2,Z_1) \hskip0.5truecm \text{Bayes Formula}
\end{eqnarray}

In fact, we are not sure this "proof" is correct as we are biased knowing that, in order to anticipate, we should do it using Bayes Formula. What allows us to get back on our feet is by introducing the Ignorance about $P(E1,E_2)$ through Eq.(\ref{eq:IgnoranceprevisionZ2C}) and Eq.(\ref{eq:IgnoranceprevisionZ2D}). However, it does not seem to be incoherent, as it is part of our Ignorance about the (futur of the) system represented by Howard.

\subsection{Case of 3 possible updates after $E_0$ : $Z_3$ state} 

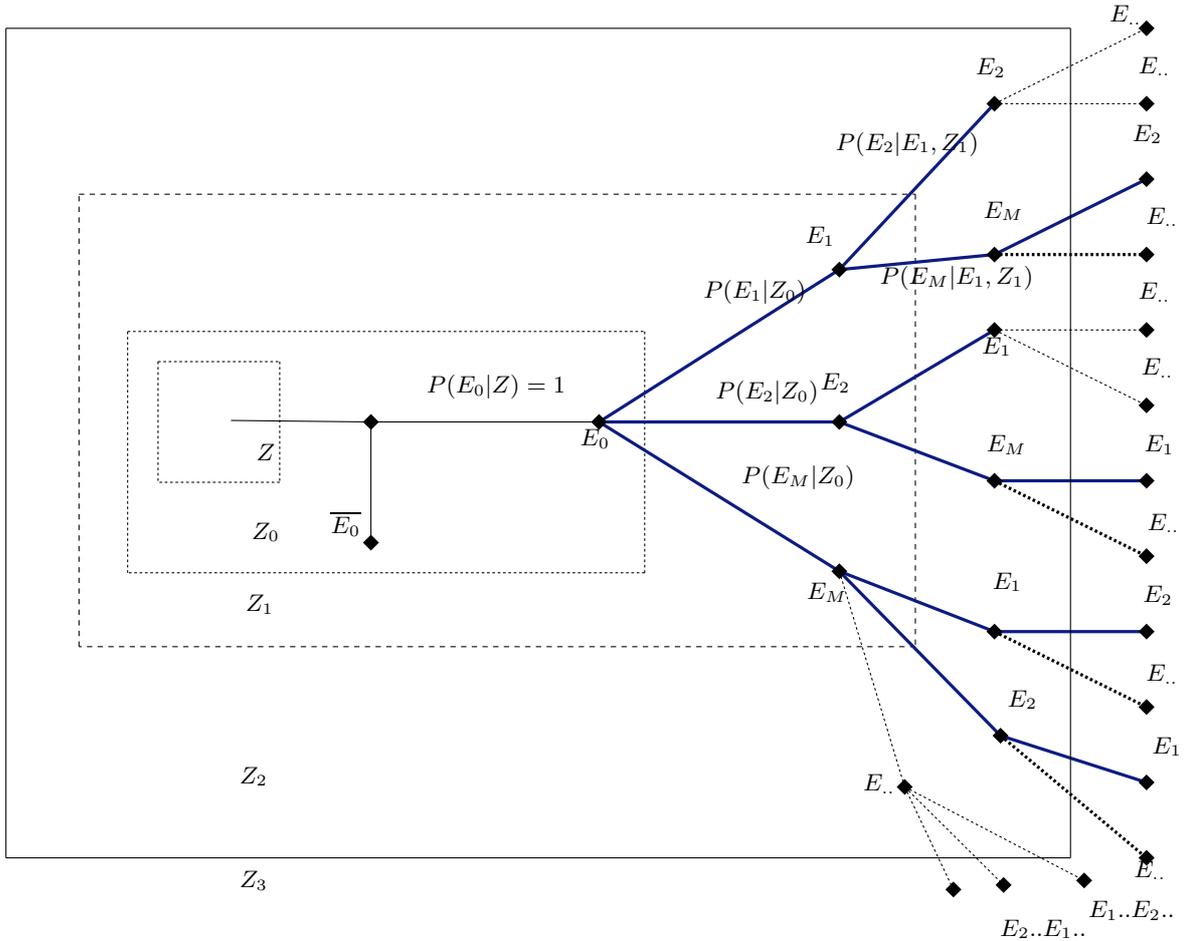
\begin{figure}[htb]
\definecolor{qqttcc}{rgb}{0.05, 0.11, 0.51}
\begin{tikzpicture}[scale=2]
\draw (8.18,-1.74)-- (6.68,-1.74);
\draw (6.99,-1.37) node[anchor=north west] {$P(E_0 | Z ) = 1$};
\draw (8,-1.73) node[anchor=north west] {$E_0$};
\draw (5.84,-2.35) node[anchor=north west] {$Z_0$};
\draw (6.68,-1.74)-- (6.68,-2.54);
\draw (6.36,-2.28) node[anchor=north west] {$\overline{E_0}$};
\draw (5.87,-1.83) node[anchor=north west] {$Z$};
\draw [line width=0.4pt,dash pattern=on 1pt off 1pt] (5.28,-1.34)-- (6.08,-1.34);
\draw [line width=0.4pt,dash pattern=on 1pt off 1pt] (6.08,-1.34)-- (6.08,-2.14);
\draw [line width=0.4pt,dash pattern=on 1pt off 1pt] (6.08,-2.14)-- (5.28,-2.14);
\draw [line width=0.4pt,dash pattern=on 1pt off 1pt] (5.28,-1.34)-- (5.28,-2.14);
\draw [dash pattern=on 1pt off 1pt] (5.08,-1.14)-- (8.48,-1.14);
\draw [dash pattern=on 1pt off 1pt] (8.48,-2.74)-- (8.48,-1.14);
\draw [dash pattern=on 1pt off 1pt] (8.48,-2.74)-- (5.08,-2.74);
\draw [dash pattern=on 1pt off 1pt] (5.08,-2.74)-- (5.08,-1.14);
\draw [line width=1.2pt,color=qqttcc] (8.18,-1.74)-- (9.76,-0.73);
\draw [line width=1.2pt,color=qqttcc] (8.18,-1.74)-- (9.76,-1.74);
\draw [line width=1.2pt,color=qqttcc] (8.18,-1.74)-- (9.76,-2.73);
\draw (9.48,-0.39) node[anchor=north west] {$E_1$};
\draw (9.58,-1.36) node[anchor=north west] {$E_2$};
\draw (9.49,-2.75) node[anchor=north west] {$E_{M}$};
\draw (8.81,-0.74) node[anchor=north west] {$P(E_1 | Z_0 )$};
\draw (8.89,-1.41) node[anchor=north west] {$P(E_2 | Z_0)$};
\draw (9.06,-1.97) node[anchor=north west] {$P(E_M | Z_0)$};
\draw (6.68,-1.74)-- (5.76,-1.73);
\draw [dash pattern=on 2pt off 2pt] (4.76,-0.23)-- (10.26,-0.23);
\draw [dash pattern=on 2pt off 2pt] (10.26,-0.23)-- (10.26,-3.23);
\draw [dash pattern=on 2pt off 2pt] (10.26,-3.23)-- (4.76,-3.23);
\draw [dash pattern=on 2pt off 2pt] (4.76,-3.23)-- (4.76,-0.23);
\draw (5.8,-2.83) node[anchor=north west] {$Z_1$};
\draw [line width=1.2pt,color=qqttcc] (9.76,-0.73)-- (10.78,0.37);
\draw [line width=1.2pt,color=qqttcc] (9.76,-0.73)-- (10.78,-0.63);
\draw [dash pattern=on 1pt off 1pt] (10.78,0.37)-- (11.78,0.37);
\draw [line width=1.2pt,dash pattern=on 1pt off 1pt] (10.78,-0.63)-- (11.78,-0.63);
\draw [dash pattern=on 1pt off 1pt] (10.78,0.37)-- (11.78,0.87);
\draw [line width=1.2pt,color=qqttcc] (10.78,-0.63)-- (11.78,-0.13);
\draw [line width=1.2pt,color=qqttcc] (9.76,-1.74)-- (10.78,-1.13);
\draw [line width=1.2pt,color=qqttcc] (9.76,-1.74)-- (10.78,-2.13);
\draw [line width=1.2pt,color=qqttcc] (9.76,-2.73)-- (10.78,-3.13);
\draw [line width=1.2pt,color=qqttcc] (9.76,-2.73)-- (10.82,-3.82);
\draw [dash pattern=on 1pt off 1pt] (10.78,-1.13)-- (11.78,-1.13);
\draw [dash pattern=on 1pt off 1pt] (10.78,-1.13)-- (11.78,-1.63);
\draw [line width=1.2pt,color=qqttcc] (10.78,-2.13)-- (11.78,-2.13);
\draw [line width=1.2pt,dash pattern=on 1pt off 1pt] (10.78,-2.13)-- (11.78,-2.63);
\draw [line width=1.2pt,color=qqttcc] (10.78,-3.13)-- (11.78,-3.13);
\draw [line width=1.2pt,dash pattern=on 1pt off 1pt] (10.78,-3.13)-- (11.78,-3.63);
\draw [line width=1.2pt,color=qqttcc] (10.82,-3.82)-- (11.78,-4.13);
\draw [line width=1.2pt,dash pattern=on 1pt off 1pt] (10.82,-3.82)-- (11.78,-4.63);
\draw (11.28,0.87)-- (4.28,0.87);
\draw (4.28,0.87)-- (4.28,-4.63);
\draw (11.28,-4.63)-- (11.28,0.87);
\draw (11.28,-4.63)-- (4.28,-4.63);
\draw (5.76,-3.97) node[anchor=north west] {$Z_2$};
\draw (5.76,-4.66) node[anchor=north west] {$Z_3$};
\draw (9.68,0.24) node[anchor=north west] {$P(E_2 | E_1, Z_1)$};
\draw (9.97,-0.65) node[anchor=north west] {$P(E_M | E_1, Z_1)$};
\draw (10.6,0.73) node[anchor=north west] {$E_2$};
\draw (11.63,0.29) node[anchor=north west] {$E_2$};
\draw (10.81,-3.46) node[anchor=north west] {$E_2$};
\draw (10.71,-2.68) node[anchor=north west] {$E_1$};
\draw (11.48,1.08) node[anchor=north west] {$E_{..}$};
\draw (11.7,-2.77) node[anchor=north west] {$E_2$};
\draw (10.65,-0.22) node[anchor=north west] {$E_{M}$};
\draw (11.67,0.74) node[anchor=north west] {$E_{..}$};
\draw (11.72,-0.26) node[anchor=north west] {$E_{..}$};
\draw (10.64,-1.12) node[anchor=north west] {$E_1$};
\draw (11.71,-1.77) node[anchor=north west] {$E_1$};
\draw (11.76,-3.77) node[anchor=north west] {$E_1$};
\draw (11.67,-0.76) node[anchor=north west] {$E_{..}$};
\draw (11.69,-1.26) node[anchor=north west] {$E_{..}$};
\draw (10.67,-1.77) node[anchor=north west] {$E_{M}$};
\draw (11.73,-2.29) node[anchor=north west] {$E_{..}$};
\draw (11.72,-3.29) node[anchor=north west] {$E_{..}$};
\draw (11.64,-4.59) node[anchor=north west] {$E_{..}$};
\draw [dash pattern=on 1pt off 1pt] (9.76,-2.73)-- (10.19,-4.16);
\draw [dash pattern=on 1pt off 1pt] (10.19,-4.16)-- (11.37,-4.78);
\draw [dash pattern=on 1pt off 1pt] (10.19,-4.16)-- (10.84,-4.81);
\draw [dash pattern=on 1pt off 1pt] (10.19,-4.16)-- (10.51,-4.84);
\draw (11.34,-4.86) node[anchor=north west] {$E_1..E_2..$};
\draw (9.86,-4.02) node[anchor=north west] {$E_{..}$};
\draw (10.76,-4.98) node[anchor=north west] {$E_2..E_1..$};
\begin{scriptsize}
\fill [color=black] (8.18,-1.74) ++(-1.5pt,0 pt) -- ++(1.5pt,1.5pt)--++(1.5pt,-1.5pt)--++(-1.5pt,-1.5pt)--++(-1.5pt,1.5pt);
\fill [color=black] (6.68,-1.74) ++(-1.5pt,0 pt) -- ++(1.5pt,1.5pt)--++(1.5pt,-1.5pt)--++(-1.5pt,-1.5pt)--++(-1.5pt,1.5pt);
\fill [color=black] (6.68,-2.54) ++(-1.5pt,0 pt) -- ++(1.5pt,1.5pt)--++(1.5pt,-1.5pt)--++(-1.5pt,-1.5pt)--++(-1.5pt,1.5pt);
\fill [color=black] (9.76,-0.73) ++(-1.5pt,0 pt) -- ++(1.5pt,1.5pt)--++(1.5pt,-1.5pt)--++(-1.5pt,-1.5pt)--++(-1.5pt,1.5pt);
\fill [color=black] (9.76,-1.74) ++(-1.5pt,0 pt) -- ++(1.5pt,1.5pt)--++(1.5pt,-1.5pt)--++(-1.5pt,-1.5pt)--++(-1.5pt,1.5pt);
\fill [color=black] (9.76,-2.73) ++(-1.5pt,0 pt) -- ++(1.5pt,1.5pt)--++(1.5pt,-1.5pt)--++(-1.5pt,-1.5pt)--++(-1.5pt,1.5pt);
\fill [color=black] (10.78,0.37) ++(-1.5pt,0 pt) -- ++(1.5pt,1.5pt)--++(1.5pt,-1.5pt)--++(-1.5pt,-1.5pt)--++(-1.5pt,1.5pt);
\fill [color=black] (10.78,-0.63) ++(-1.5pt,0 pt) -- ++(1.5pt,1.5pt)--++(1.5pt,-1.5pt)--++(-1.5pt,-1.5pt)--++(-1.5pt,1.5pt);
\fill [color=black] (11.78,0.37) ++(-1.5pt,0 pt) -- ++(1.5pt,1.5pt)--++(1.5pt,-1.5pt)--++(-1.5pt,-1.5pt)--++(-1.5pt,1.5pt);
\fill [color=black] (11.78,-0.63) ++(-1.5pt,0 pt) -- ++(1.5pt,1.5pt)--++(1.5pt,-1.5pt)--++(-1.5pt,-1.5pt)--++(-1.5pt,1.5pt);
\fill [color=black] (11.78,0.87) ++(-1.5pt,0 pt) -- ++(1.5pt,1.5pt)--++(1.5pt,-1.5pt)--++(-1.5pt,-1.5pt)--++(-1.5pt,1.5pt);
\fill [color=black] (11.78,-0.13) ++(-1.5pt,0 pt) -- ++(1.5pt,1.5pt)--++(1.5pt,-1.5pt)--++(-1.5pt,-1.5pt)--++(-1.5pt,1.5pt);
\fill [color=black] (10.78,-1.13) ++(-1.5pt,0 pt) -- ++(1.5pt,1.5pt)--++(1.5pt,-1.5pt)--++(-1.5pt,-1.5pt)--++(-1.5pt,1.5pt);
\fill [color=black] (10.78,-2.13) ++(-1.5pt,0 pt) -- ++(1.5pt,1.5pt)--++(1.5pt,-1.5pt)--++(-1.5pt,-1.5pt)--++(-1.5pt,1.5pt);
\fill [color=black] (10.78,-3.13) ++(-1.5pt,0 pt) -- ++(1.5pt,1.5pt)--++(1.5pt,-1.5pt)--++(-1.5pt,-1.5pt)--++(-1.5pt,1.5pt);
\fill [color=black] (10.82,-3.82) ++(-1.5pt,0 pt) -- ++(1.5pt,1.5pt)--++(1.5pt,-1.5pt)--++(-1.5pt,-1.5pt)--++(-1.5pt,1.5pt);
\fill [color=black] (11.78,-1.13) ++(-1.5pt,0 pt) -- ++(1.5pt,1.5pt)--++(1.5pt,-1.5pt)--++(-1.5pt,-1.5pt)--++(-1.5pt,1.5pt);
\fill [color=black] (11.78,-1.63) ++(-1.5pt,0 pt) -- ++(1.5pt,1.5pt)--++(1.5pt,-1.5pt)--++(-1.5pt,-1.5pt)--++(-1.5pt,1.5pt);
\fill [color=black] (11.78,-2.13) ++(-1.5pt,0 pt) -- ++(1.5pt,1.5pt)--++(1.5pt,-1.5pt)--++(-1.5pt,-1.5pt)--++(-1.5pt,1.5pt);
\fill [color=black] (11.78,-2.63) ++(-1.5pt,0 pt) -- ++(1.5pt,1.5pt)--++(1.5pt,-1.5pt)--++(-1.5pt,-1.5pt)--++(-1.5pt,1.5pt);
\fill [color=black] (11.78,-3.13) ++(-1.5pt,0 pt) -- ++(1.5pt,1.5pt)--++(1.5pt,-1.5pt)--++(-1.5pt,-1.5pt)--++(-1.5pt,1.5pt);
\fill [color=black] (11.78,-3.63) ++(-1.5pt,0 pt) -- ++(1.5pt,1.5pt)--++(1.5pt,-1.5pt)--++(-1.5pt,-1.5pt)--++(-1.5pt,1.5pt);
\fill [color=black] (11.78,-4.13) ++(-1.5pt,0 pt) -- ++(1.5pt,1.5pt)--++(1.5pt,-1.5pt)--++(-1.5pt,-1.5pt)--++(-1.5pt,1.5pt);
\fill [color=black] (11.78,-4.63) ++(-1.5pt,0 pt) -- ++(1.5pt,1.5pt)--++(1.5pt,-1.5pt)--++(-1.5pt,-1.5pt)--++(-1.5pt,1.5pt);
\fill [color=black] (10.19,-4.16) ++(-1.5pt,0 pt) -- ++(1.5pt,1.5pt)--++(1.5pt,-1.5pt)--++(-1.5pt,-1.5pt)--++(-1.5pt,1.5pt);
\fill [color=black] (11.37,-4.78) ++(-1.5pt,0 pt) -- ++(1.5pt,1.5pt)--++(1.5pt,-1.5pt)--++(-1.5pt,-1.5pt)--++(-1.5pt,1.5pt);
\fill [color=black] (10.84,-4.81) ++(-1.5pt,0 pt) -- ++(1.5pt,1.5pt)--++(1.5pt,-1.5pt)--++(-1.5pt,-1.5pt)--++(-1.5pt,1.5pt);
\fill [color=black] (10.51,-4.84) ++(-1.5pt,0 pt) -- ++(1.5pt,1.5pt)--++(1.5pt,-1.5pt)--++(-1.5pt,-1.5pt)--++(-1.5pt,1.5pt);
\end{scriptsize}
\end{tikzpicture}
\caption{Case of 3 possible updates after $E_0$} 
\label{fig:3updatesafterE0}
\end{figure}

With Fig.(\ref{fig:3updatesafterE0}), we are wondering about what is the more general Bayes Formula where we should anticipate so much more than 2 steps ahead. We start by first wondering if 
\begin{equation}
P(E_1,E_2 |Z_2) = \displaystyle \sum_i^6  \,\, k_i \,\,P( \{\cap, E_1, E_2,E_{..} \} |Z_2) 
\end{equation} 
where $\{\cap, E_1, E_2,E_{..} \}$ represent the 6 permutations of $E_1$, $E_2$, $E_{..}$. By doing as previously, we would think that the derivatives of the Ignorance should lead also to 
\begin{eqnarray}
& & P(E_1\cap E_2) = P(E_2 \cap E_1) \\
&\Leftrightarrow & \displaystyle \sum_i^3 P(1 \rightarrow 2 ,E_{..}) = \sum_i^3 P(2 \rightarrow 1 ,E_{..}) \\
&\Leftrightarrow & P(E_{..} \cap E_1 \cap E_2)  + P(E_1 \cap E_{..} \cap E_2)  + P(E_1\cap E_2 \cap E_{..} ) = P(E_{..} \cap E_2 \cap E_1)  + P(E_2 \cap E_{..} \cap E_1)  + P(E_2\cap E_1 \cap E_{..} )  \nonumber 
\end{eqnarray}
where $P(i \rightarrow j ,E_{..})$ is such that $E_i$ appears always before $E_j$.\sk 
In particular, as the case $E_{..} \rightarrow (E_1 , E_2) $ where $E_1$ and $E_2$ would appear at the next update (the last two branches at the bottom in our graph), is in fact the same situation as here where $E_{..}$ would be considered as $E_0$, we would say that 
\begin{itemize}
\item $P(E_{..}\cap E_1 \cap E_2) = P(E_{..}\cap E_2 \cap E_1)$,
\item but also that $P(E_1 \cap E_{..} \cap E_2) = P(E_2 \cap E_{..} \cap E_1)$
\item and thus $P(E_1\cap E_2 \cap E_{..} ) = P(E_2 \cap E_1 \cap E_{..})$. 
\end{itemize}
This would seem coherent even for the general case, that is, the Bayes formula $P( \{i, j\} , ..) =  P( \{j, i\} , ..)$ works for every configurations with symmetry $i \leftrightarrow j$, as the 3 equalities just before.  \sk 
This is of course not a rigorous mathematical proof, and we leave that for later if necessary or if it has not been demonstrated somewhere else (as it should not be new).

\subsection{Ignoring the anticipation is Ignoring Bayes Formula, that is the question} 

We have shown here that including probabilities on the states where $E_1$ and $E_2$ would appear during its journey, thus the states where $E_1 \cap E_2$ and $E_2 \cap E_1$, leads us to incorporate in our Ignorance new terms which account for the fact that we do not also know the probabilities of the joined events, which are however also calculable \textit{in fine}, with the basic probabilities of the tree ! Consequently, by minimizing Ignorance, we (think that we) were able to recover Bayes Formula as a deep tool to anticipate, knowing some caracteristics, what would  be the values of the other ones.\sk 

That is to say otherwise that, including in the Ignorance all the probabilities one can construct from the probabilities of the tree, this unique quantity representing Howard's Ignorance, due to its complexity where probabilities are linked, will encode all what one should know and guess about the situation. Therefore all informations about what Howard knows, ignores, can guess or not, could be condensed in just the Ignorance, representing his memory, and the mecanism to remember or anticipate his knowledge could be the one that minimize his Ignorance, respectively with respect to constraints for memories, and to the probabilities for the two kind of anticipations (the one about his current situation, and the one one step ahead).

\section{Two sides of a coin lead to a Duck} 

\subsection{The algorithm of an update}

Opening the door, Howard finds himself in front of a new corridor in the middle of which, on the floor, an object still unknown to Howard shines. How does it translate into Howard's brain ?  
Firstly, in his "tree of knowledge", Howard is now on the branch $\star$ on Fig.(\ref{fig:2updatesafterE0}), where $E_M$ ("something else") is verified, and therefore leads to $(E_1,E_2,E_ {..})$. Secondly, in his brain, 
\begin{itemize}
\item An event occurs and his state of knwoledge is updated $Z_1 \rightarrow Z_2(Z_1,x^\mu)$,
\item $E_M$ is becoming known such as $E_M \rightarrow E_a$ "I have met an object", and thus, a new constraint appears $0 = P(E_a |Z_1)-1$ in his Ignorance, and so, in his memory.
\item Consequently, a new branch $E_M$ is created due to the update, that represents again what he does not know yet. Howard is now facing 3 possibilities $(E_1,E_2,E_M)$ and the probabilities are also updated as they respect the constraint $0 = P(E_1|Z_1) + P(E_2|Z_1) + P(E_M|Z_1) -1$  (same as before but with the update $Z_1 \rightarrow Z_2$),
\item his "basic" Ignorance is now 
\begin{eqnarray}
H[.|Z_2] &=& \mu_0 (P(E_0|Z_1)-1) + \mu_a (P(E_a|Z_1)-1)  + \mu_1 (P(E_1|Z_1) + P(E_2|Z_1) + P(E_M|Z_1) -1) \label{eq:ignoranceIci_coin_memoire}\\
&& - \displaystyle \sum_{i =1,2,M} \lambda_i P(E_i|Z_1) \ln P(E_i|Z_1)\label{eq:ignoranceIci_coin_incertitudes}
\end{eqnarray}
where Eq.(\ref{eq:ignoranceIci_coin_memoire}) and Eq.(\ref{eq:ignoranceIci_coin_incertitudes}) represents respectively what he knows and so what is in his memory, and his uncertainties. In the following, except if we say that he forgets one specific knowledge (that is putting $\mu_i = 0$), we will write $\Sigma$ in Eq.(\ref{eq:ignoranceIci_coin_memoire}) to indicate what is learned.
\end{itemize}

\subsection{What could it be ? } 
He could also anticipate and update his \textit{a priori}, that is, his \textit{credences} which mesure his belief strength about the nature of the object. He could then create three new propositions : knowing that $E_1$ and $E_2$ are possible, 
\begin{enumerate}
\item "the object is a coin" : $E_{c}$ 
\item "the object is a dice" : $E_{d}$ 
\item "the object is a something else" : $E_{s}$ 
\end{enumerate}
How would it translate in the tree of probabilities ? Let us think. \sk 
As the propositions above are consequences of $E_1$ and $E_2$, their probabilities would be of the form $P(E_i | E_1,E_2,Z_1)$ and therefore one would know, first, for instance $E_1$, that a coin exists, before assuming that $E_c$ is true. However, this does not mean that one can not verify $E_c$ before $E_1$ : indeed, if Howard did not have the knowledge $E_1$ that a coin exist before coming across the object, and if a paper had the proposition $E_c$ written on it, by reading it from left to right, Howard would know that, first, this is related to the object in the corridor, and secondly, that it is a coin, knowing therefore that, according to the situation, a coin exists, verifying that $E_1$ is true as a direct consequence.\sk 
Comment : can he be sure, certain, that this object is really a coin ? The answer is logically no as he has no way to assert that outside his current situation, but for him, he can consider without trouble that it is true, as it is just putting a word without clear concepts behind, on a representation of an object. Knowing $E_c$ is, at this point, true, his memory about it would have the expression $P(E_c|Z_n) -1 = 0$ with a Lagrange multiplier which will be zero if he forgets about this, or he can transforme the proposition if somehow later, he is able to know that the object was not in fact a coin : for instance, if living with a lot of persons, all of them call it a "duck", then he should think that, statiscally, he should follow the main opinion about it and change his memory by including terms like 
\begin{equation}
H_new[E_c,.] = \alpha(P(E_c|Z_n)-0) + \beta (P(\text{"this object is a duck"}|Z_n) -1) + \gamma(P("\text{I was wrong at }Z_{n-1}" |Z_n)-1) \rightarrow 0
\end{equation}
This leads us to consider the following tree 
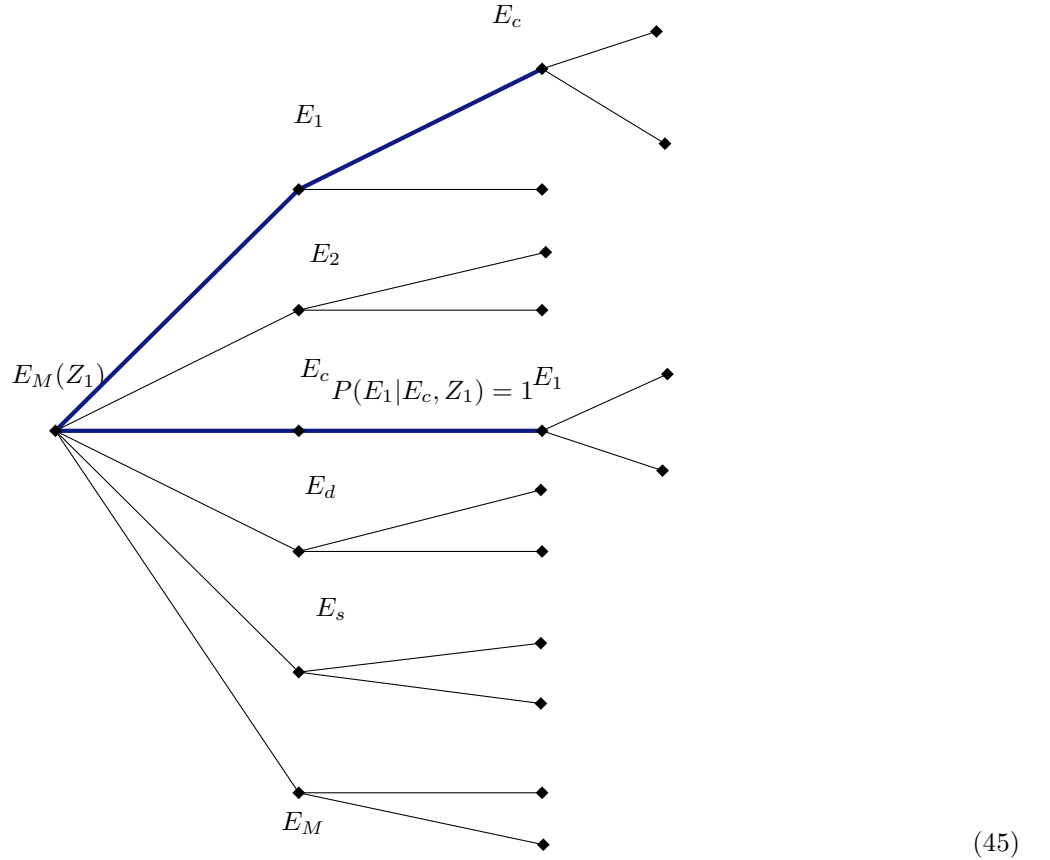
\begin{equation}\label{Graph:coincoin_2}
\definecolor{qqqqcc}{rgb}{0.05, 0.11, 0.51}
\begin{tikzpicture}[scale=1.6]
\draw (-1.28,0.42) node[anchor=north west] {$E_{M} (Z_1)$};
\draw [line width=1.6pt,color=qqqqcc] (-0.84,-0.22)-- (1.16,1.78);
\draw [line width=1.6pt,color=qqqqcc] (-0.84,-0.22)-- (1.16,-0.22);
\draw (-0.84,-0.22)-- (1.16,-1.22);
\draw (-0.84,-0.22)-- (1.16,0.78);
\draw (-0.84,-0.22)-- (1.16,-2.22);
\draw (-0.84,-0.22)-- (1.16,-3.22);
\draw (1.04,2.56) node[anchor=north west] {$E_1$};
\draw (1.17,1.41) node[anchor=north west] {$E_2$};
\draw (1.09,0.45) node[anchor=north west] {$E_c$};
\draw (1.13,-0.52) node[anchor=north west] {$E_d$};
\draw (1.22,-1.53) node[anchor=north west] {$E_s$};
\draw (0.94,-3.3) node[anchor=north west] {$E_{M}$};
\draw (3.16,-0.22)-- (4.19,0.25);
\draw (3,0.39) node[anchor=north west] {$E_1$};
\draw [line width=1.6pt,color=qqqqcc] (1.16,1.78)-- (3.16,2.78);
\draw (1.16,1.78)-- (3.16,1.78);
\draw (3.16,-0.22)-- (4.15,-0.55);
\draw [line width=1.6pt,color=qqqqcc] (1.16,-0.22)-- (3.16,-0.22);
\draw (2.67,3.39) node[anchor=north west] {$E_c$};
\draw (4.1,3.09)-- (3.16,2.78);
\draw (3.16,2.78)-- (4.17,2.16);
\draw (1.16,0.78)-- (3.19,1.26);
\draw (1.16,0.78)-- (3.16,0.78);
\draw (1.35,0.29) node[anchor=north west] {$P(E_1 | E_c, Z_1) = 1$};
\draw (1.16,-1.22)-- (3.15,-0.71);
\draw (1.16,-1.22)-- (3.16,-1.22);
\draw (1.16,-2.22)-- (3.15,-1.98);
\draw (1.16,-3.22)-- (3.16,-3.22);
\draw (1.16,-3.22)-- (3.17,-3.65);
\draw (3.15,-2.48)-- (1.16,-2.22);
\begin{scriptsize}
\fill [color=black] (-0.84,-0.22) ++(-1.5pt,0 pt) -- ++(1.5pt,1.5pt)--++(1.5pt,-1.5pt)--++(-1.5pt,-1.5pt)--++(-1.5pt,1.5pt);
\fill [color=black] (1.16,1.78) ++(-1.5pt,0 pt) -- ++(1.5pt,1.5pt)--++(1.5pt,-1.5pt)--++(-1.5pt,-1.5pt)--++(-1.5pt,1.5pt);
\fill [color=black] (1.16,-0.22) ++(-1.5pt,0 pt) -- ++(1.5pt,1.5pt)--++(1.5pt,-1.5pt)--++(-1.5pt,-1.5pt)--++(-1.5pt,1.5pt);
\fill [color=black] (1.16,-1.22) ++(-1.5pt,0 pt) -- ++(1.5pt,1.5pt)--++(1.5pt,-1.5pt)--++(-1.5pt,-1.5pt)--++(-1.5pt,1.5pt);
\fill [color=black] (1.16,0.78) ++(-1.5pt,0 pt) -- ++(1.5pt,1.5pt)--++(1.5pt,-1.5pt)--++(-1.5pt,-1.5pt)--++(-1.5pt,1.5pt);
\fill [color=black] (1.16,-2.22) ++(-1.5pt,0 pt) -- ++(1.5pt,1.5pt)--++(1.5pt,-1.5pt)--++(-1.5pt,-1.5pt)--++(-1.5pt,1.5pt);
\fill [color=black] (1.16,-3.22) ++(-1.5pt,0 pt) -- ++(1.5pt,1.5pt)--++(1.5pt,-1.5pt)--++(-1.5pt,-1.5pt)--++(-1.5pt,1.5pt);
\fill [color=black] (3.16,-0.22) ++(-1.5pt,0 pt) -- ++(1.5pt,1.5pt)--++(1.5pt,-1.5pt)--++(-1.5pt,-1.5pt)--++(-1.5pt,1.5pt);
\fill [color=black] (4.19,0.25) ++(-1.5pt,0 pt) -- ++(1.5pt,1.5pt)--++(1.5pt,-1.5pt)--++(-1.5pt,-1.5pt)--++(-1.5pt,1.5pt);
\fill [color=black] (3.16,2.78) ++(-1.5pt,0 pt) -- ++(1.5pt,1.5pt)--++(1.5pt,-1.5pt)--++(-1.5pt,-1.5pt)--++(-1.5pt,1.5pt);
\fill [color=black] (3.16,1.78) ++(-1.5pt,0 pt) -- ++(1.5pt,1.5pt)--++(1.5pt,-1.5pt)--++(-1.5pt,-1.5pt)--++(-1.5pt,1.5pt);
\fill [color=black] (4.15,-0.55) ++(-1.5pt,0 pt) -- ++(1.5pt,1.5pt)--++(1.5pt,-1.5pt)--++(-1.5pt,-1.5pt)--++(-1.5pt,1.5pt);
\fill [color=black] (4.1,3.09) ++(-1.5pt,0 pt) -- ++(1.5pt,1.5pt)--++(1.5pt,-1.5pt)--++(-1.5pt,-1.5pt)--++(-1.5pt,1.5pt);
\fill [color=black] (4.17,2.16) ++(-1.5pt,0 pt) -- ++(1.5pt,1.5pt)--++(1.5pt,-1.5pt)--++(-1.5pt,-1.5pt)--++(-1.5pt,1.5pt);
\fill [color=black] (3.19,1.26) ++(-1.5pt,0 pt) -- ++(1.5pt,1.5pt)--++(1.5pt,-1.5pt)--++(-1.5pt,-1.5pt)--++(-1.5pt,1.5pt);
\fill [color=black] (3.16,0.78) ++(-1.5pt,0 pt) -- ++(1.5pt,1.5pt)--++(1.5pt,-1.5pt)--++(-1.5pt,-1.5pt)--++(-1.5pt,1.5pt);
\fill [color=black] (3.16,-1.22) ++(-1.5pt,0 pt) -- ++(1.5pt,1.5pt)--++(1.5pt,-1.5pt)--++(-1.5pt,-1.5pt)--++(-1.5pt,1.5pt);
\fill [color=black] (3.15,-0.71) ++(-1.5pt,0 pt) -- ++(1.5pt,1.5pt)--++(1.5pt,-1.5pt)--++(-1.5pt,-1.5pt)--++(-1.5pt,1.5pt);
\fill [color=black] (3.15,-1.98) ++(-1.5pt,0 pt) -- ++(1.5pt,1.5pt)--++(1.5pt,-1.5pt)--++(-1.5pt,-1.5pt)--++(-1.5pt,1.5pt);
\fill [color=black] (3.15,-2.48) ++(-1.5pt,0 pt) -- ++(1.5pt,1.5pt)--++(1.5pt,-1.5pt)--++(-1.5pt,-1.5pt)--++(-1.5pt,1.5pt);
\fill [color=black] (3.16,-3.22) ++(-1.5pt,0 pt) -- ++(1.5pt,1.5pt)--++(1.5pt,-1.5pt)--++(-1.5pt,-1.5pt)--++(-1.5pt,1.5pt);
\fill [color=black] (3.17,-3.65) ++(-1.5pt,0 pt) -- ++(1.5pt,1.5pt)--++(1.5pt,-1.5pt)--++(-1.5pt,-1.5pt)--++(-1.5pt,1.5pt);
\end{scriptsize}
\end{tikzpicture}
\end{equation}
Then, by saying that events are all equiprobable (he is new to this world and therefore has no experience), he could think that 
\begin{itemize}
\item $P(E_1|Z_1) = P(E_2|Z_1) = P(E_c|Z_1) = P(E_d|Z_1) = P(E_s|Z_1)  = P(E_M|Z_1) = \dfrac{1}{6}$
\item but also that $P(E_c|E_1,Z_2) = \dfrac{1}{5}$. 
\end{itemize}
\pagebreak
However, because $P(E_1|E_c) = 1$, one would have 
\begin{itemize}
\item $P(E_1 \cap E_c) = P(E_1) \times P(E_c|E_1) = \dfrac{1}{6} \times \dfrac{1}{5} = \dfrac{1}{30}$
\item $P(E_c \cap E_1) = P(E_c) \times P(E_1|E_c) = \dfrac{1}{6} \times 1 = \dfrac{1}{6}$
\end{itemize}
One reason that these two quantities are not equal, assuming that they should be as it seems logical, is due to the branches $E_M \rightarrow E_M$ and $E_1 \rightarrow E_M$ (not represented on Fig.(\ref{Graph:coincoin_2})) where Howard supposed that their weight is the same as all of the others but he does not know if there is only one event or an infinty in $E_M$ (as shown in the algorithm before). Nevertheless, another strange thing seems to exist : from Bayes formula, we can see that 
\begin{equation}
P(E_1 \cap E_c) = P(E_c \cap E_1) \Leftrightarrow P(E_1) \times P(E_c|E_1) = P(E_c) \times 1 = P(E_c) 
\end{equation}
which could have different results 
\begin{itemize}
\item  $P(E_1) = P(E_c)  = 0$ there are no coins, fine
\item  $P(E_c|E_1) = P(E_c)  = 0$ this object is not a coin, fine
\item $P(E_1) = P(E_c) $  not null, and therefore $P(E_c|E_1) = 1$ : if we find an object, we will think that this object is surely a coin. In fact this is true at least when $P(E_c) = 1$ that is when we verify that the object is really a coin, implying that $P(E_1) = 1$ and so $P(E_c|E_1) = 1$. 
\item $P(E_c|E_1) = \dfrac{P(E_c)}{P(E_1)} \leq 1 $ , with $P(E_1) >0$ and $P(E_c)\leq P(E_1)$. In the equiprobable case, then $P(E_c) = P(E_1)$ and so we recover the case before, even if the number of branches at one node goes to infinity.
\item $P(E_1) = 1$ and so $P(E_c|E_1) = P(E_c)$, that is, if we verify that a coin exists, then from this assertion, we do not know if this object is a coin : both events are independent.
\end{itemize}
This would be also true for $E_2$ and $E_d$. However, we do not know yet if, due to the complexity of the tree, where $E_c$ and $E_1$ can appear in many other branches, our interpretation should be a lot more complexe.

\subsection{Here comes Nyarlathotep}\label{sec:here_comes_Nyarlathotep}

Not knowing anything else, Howard was not able to decide about the nature of the object, therefore he set in his mind that $P(E_c|Z_1) = P(E_d|Z_1) = P(E_s|Z_1)$. In this case, his Ignorance could be written as 
\begin{eqnarray}
H[.|Z_2] &=& \Sigma +  \mu_1 (P(E_1|Z_1) + P(E_2|Z_1) + P(E_c|Z_1) + P(E_d|Z_1) + P(E_s|Z_1) + P(E_M|Z_1) - 1) \\
&& + \alpha_1 (P(E_c|Z_1) - P(E_d|Z_1)) + \alpha_2 (P(E_d|Z_1) - P(E_s|Z_1)) + \alpha_3 (P(E_s|Z_1) - P(E_c|Z_1)) \\
&& - \displaystyle \sum_i \lambda_i P(E_i |Z_1) \ln P(E_i |Z_1)
\end{eqnarray}
Derivatives w.r.t $P(E_i |Z_1)$ for $i \in \{c,d,s\}$ give 
\begin{equation}
\left.\dfrac{\delta H}{\delta P(E_c |Z_1)}\right|_{Z_2}= 0 \Leftrightarrow  \mu_1 + \alpha_1-\alpha_3 - \lambda_c (\ln P(E_c |Z_1) + 1) = 0
\end{equation}
\begin{equation}
\left.\dfrac{\delta H}{\delta P(E_d |Z_1)}\right|_{Z_2}= 0 \Leftrightarrow  \mu_1 + \alpha_2-\alpha_1 - \lambda_d (\ln P(E_d |Z_1) + 1) = 0
\end{equation}
\begin{equation}
\left.\dfrac{\delta H}{\delta P(E_s |Z_1)}\right|_{Z_2} = 0 \Leftrightarrow  \mu_1 + \alpha_3-\alpha_2 - \lambda_s (\ln P(E_s |Z_1) + 1) = 0
\end{equation}
However, as they concerne all the object, we could set $\alpha_1 = \alpha_2 = \alpha_3 = \alpha$ and $\lambda_i = \lambda$ for $i\in \{c,d,s\}$, leading to the same probability such that $\mu1  = \lambda (\ln(P(E_i|Z_2) + 1)$ which is in fact the same probability as $P(E_1|Z_2)$, $P(E_2|Z_2)$ and $P(E_M|Z_2)$. \sk
Adding constraints, in this case, does not bring anything new. The reason is that the constraints he set are not "real" in the sens that they are coming from Howard's opinion and concern therefore his anticipation. Moreover, as shown before, the result should be fine because, setting $\mu = \lambda$ leads again to $P(E_i|Z_2) = 1$ : all events have the same probability to appear to Howard from the point of view of Howard, and the one which would appear is the one ... which appears. This is not a system which send results like a dice or a coin, Howard is a system which receives interactions from outside and so far, its environment does not favour one piece of information over another (he is not closed to a particular source of information that would predominate in Howard's data acquisition). Thus, one should in general not think like that, at least maybe not setting for instance the constraints as having all the same value. So, how constraints could influence Howard ? Here comes Nyarlathotep. 

\subsubsection{What does it change ?}

Hidden from Howard's view, Nyarlathotep watches him. Determined to influence Howard's fate, at the speed of light squared, he substitutes the object that Howard was observing, which was a coin, puts it in a bag containing six other coins, two dice and a duck, and, randomly and without looking, scatters them in each corridor around him, including one object in the one where Howard is standing. Nyarlathotep has also taken care to place a piece of paper underneath each object, stating "this object is a ...". Making it so that Howard's next action is to pick up the object, read the paper and check one of the propositions $E_c$, $E_d$ or $E_s$. The probability that Howard observes a coin is now $P(E_c|Z_1) = 0.7$, and is the most likely. This knowledge reflects the situation in which Howard is, but seen from the point of view of Nyarlathotep. \sk 

Nyarlathotep then decides to reveal himself, opens the door and looking Howard in the eye, he points his finger at the object and says "My name is Nobody, the probability that this object is a coin is 70\%". Not surprised at all, Howard update his knowledge $ Z_2 \rightarrow Z_3$, setting $E_N$ "I have met Nobody" and according to him, $P(E_c|Z_3(E_N))-0.7 = 0$. Of course, for Howard, this is an opinion, but as it is also representing the situation, Howard considers it as another constraint (we could say that Nyarlathotep tricks his brain to do as such).  How to translate it with the Ignorance ? \sk 

We could add a constraint as before such that 
\begin{eqnarray}
H[.|Z_3] &=& \Sigma +  \mu_1 (P(E_1|Z_2) + P(E_2|Z_2) + P(E_c|Z_2) + P(E_d|Z_2) + P(E_s|Z_2) + P(E_M|Z_2) - 1) \\
&& + \alpha_1 (P(E_c|Z_2) - 0.7) \\
&& - \displaystyle \sum_i \lambda_i P(E_i |Z_2) \ln P(E_i |Z_2)
\end{eqnarray}
and therefore 
\begin{equation}
\left.\dfrac{\delta H[.|Z_3]}{\delta P(E_c|Z_2)}\right|_{Z_3} = 0  \Leftrightarrow \mu_1 + \alpha_1 - \lambda_c (\ln P(E_c|Z_2) +1) = 0
\end{equation}
So, how appears the 0.7 as it should play a role ? We could have also set $\alpha_2(P(E_d|Z_2) -0.2)$ and again, "0.2" would not appear. What we could do is to put the information about 0.7 such that at the end $P(E_c|Z_3) = 0.7$ ... as an anticipation ! (again, if it is a coin, then for Howard it will be $P(E_c|Z_3) = 1$ when he will assert that). Therefore, setting $\mu_1 = \lambda_i$ for all $i$, we would solve 
\begin{equation}
\left.\dfrac{\delta H[.|Z_3]}{\delta P(E_c|Z_2)}\right|_{Z_3} = 0  \Leftrightarrow  \mu + \alpha_1 - \mu (\ln (0.7) +1) = 0 \Leftrightarrow  \alpha_1 = \mu \ln(0.7)
\end{equation}
which is the same as setting the Ignorance as 
\begin{eqnarray}
H[.|Z_3] &=& \Sigma +  \mu (P(E_1|Z_2) + P(E_2|Z_2) + P(E_d|Z_2) + P(E_s|Z_2) + P(E_M|Z_2) - 1) \\
&& + \mu (1 + \ln(0.7)) P(E_c|Z_2)\\
&& - \displaystyle \sum_i \mu P(E_i |Z_2) \ln P(E_i |Z_2)
\end{eqnarray}
that is for $P(E_c|Z_2)$ : $\mu P(E_c|Z_2) (\ln(0.7) - \ln P(E_c|Z_2)) = \mu P(E_c|Z_2) \ln \left( \dfrac{0.7}{P(E_c|Z_2)} \right)$. However, this does not work because doing so ...
\begin{equation}
\left.\dfrac{\delta H[.|Z_3]}{\delta P(E_i|Z_2)}\right|_{i\neq c, Z_3} = 0  \Leftrightarrow  P(E_i|Z_3) = 1 > P(E_c|Z_3) 
\end{equation}
therefore $P(E_c|Z_2)$ is the most probable but not according to the minimization of Ignorance.\sk
Howard could also add another proposition to $E_N$ : if we set $E_X$ : "the probability of $P(E_c|Z_3(E_N))$ is 0.7", this would be verified by Howard only in a multiverse, after a thousand and thousand of attempts. Therefore, it is not considered as relevant here. \sk 
A possible solution was given a century ago by A. Einstein who could have said "Everything is described under a general and special relativity". Of course, we are not talking about the description of the dynamical evolutions of objects including space-time, nor the description of objects in quantum domains by Carlo Rovelli's theory of relational quantum mechanics \cite{Rovelli} (for which the use of Shannon ignorance or constrained entropy might be useful), but as everything is relatively described from one to another, so are the probabilities ! If we "want" to keep notion about the percentage of a probability, we have to think as such ! What represents $P(E_c|Z_2) = 0.7$ in our case ? It means that $P(E_c|Z_3)$ takes 70\% of the total probability, that is $P(E_c|Z_3) = 0.7 (P(E_1|Z_2) + P(E_2|Z_2) + P(E_c|Z_2) + P(E_d|Z_2) + P(E_s|Z_2) + P(E_M|Z_2) )$ and so the constraint would become.
\begin{eqnarray}
& & P(E_c|Z_3) - 0.7 (P(E_1|Z_2) + P(E_2|Z_2) + P(E_c|Z_2) + P(E_d|Z_2) + P(E_s|Z_2) + P(E_M|Z_2) ) = 0  \\ 
&\Leftrightarrow & 0.3  P(E_c|Z_2)- 0.7 (P(E_1|Z_2) + P(E_2|Z_2) + P(E_d|Z_2) + P(E_s|Z_2) + P(E_M|Z_2) )  = 0
\end{eqnarray}

\newpage

Keeping a more general case where we set $P(E_c|Z_2) = \beta$, with $0\leq \beta \leq 1$, the constraint is now
\begin{equation}\label{eq:constraints_beta_NN}
(1 - \beta)  P(E_c|Z_2)- \beta  (P(E_1|Z_2) + P(E_2|Z_2) + P(E_d|Z_2) + P(E_s|Z_2) + P(E_M|Z_2) )  = 0
\end{equation}
and the Ignorance is 
\begin{eqnarray}
H[.|Z_3] &=& \Sigma + \mu \left[\dfrac{{}^{}}{{}^{}} P(E_1|Z_2) + P(E_2|Z_2) + P(E_c|Z_2) + P(E_d|Z_2) + P(E_s|Z_2) + P(E_M|Z_2)  -1 \right] \\
&& + \alpha \left[\dfrac{{}^{}}{{}^{}} (1 - \beta)  P(E_c|Z_2)- \beta  (P(E_1|Z_2) + P(E_2|Z_2) + P(E_d|Z_2) + P(E_s|Z_2) + P(E_M|Z_2) )   \right] \\
&& - \displaystyle \sum_i \lambda_i P(E_i|Z_2) \ln P(E_i|Z_2).
\end{eqnarray}

Derivatives with respect to $P(E_i|Z_2)$ for $i\neq c$ leads to 
\begin{eqnarray}
\left. \dfrac{\delta H[.|Z_2]}{\delta P(E_i|Z_2)}\right|_{Z_3} = 0 & \Leftrightarrow & -\lambda_i (\ln P(E_i|Z_3)+1) + \mu - \alpha \beta = 0 \\
& \Leftrightarrow &  \mu - \alpha \beta =\lambda_i (\ln P(E_i|Z_3)+1) \\
& \Leftrightarrow &  \dfrac{\mu - \alpha \beta}{\lambda_i}-1  =\ln P(E_i|Z_3)  \label{eq:expressionP(E_i|Z_3)_a} \\
& \Leftrightarrow &  P(E_i|Z_3) = exp \left(\dfrac{\mu}{\lambda_i} - \dfrac{\alpha \beta}{\lambda_i} -1 \right) \,\,\,\,  \xrightarrow[]{\mu = \lambda_i}  \,\,\,\, P(E_i|Z_3) = e^{-\dfrac{\alpha \beta}{\mu}}. \label{eq:expressionP(E_i|Z_3)_b}
\end{eqnarray}

Derivatives with respect to $P(E_c|Z_2)$ leads to  
\begin{eqnarray}
\left. \dfrac{\delta H[.|Z_2]}{\delta P(E_c|Z_2)}\right|_{Z_3} = 0 & \Leftrightarrow & -\lambda_c (\ln P(E_c|Z_3)+1) + \mu + \alpha(1- \beta) = 0 \\
& \Leftrightarrow &  \mu - \alpha \beta =\lambda_c (\ln P(E_c|Z_3)+1) - \alpha \\
& \Leftrightarrow &  \dfrac{\mu + \alpha (1-\beta)}{\lambda_c} -1  =\ln P(E_c|Z_3) \label{eq:expressionP(E_c|Z_3)_a} \\
& \Leftrightarrow &  P(E_c|Z_3) = exp \left(\dfrac{\alpha}{\lambda_c} + \dfrac{\mu}{\lambda_c}  - \dfrac{\alpha \beta}{\lambda_c} -1 \right) \,\,\,\,  \xrightarrow[]{\mu = \lambda_i}  \,\,\,\, P(E_c|Z_3) = e^{\dfrac{\alpha (1-\beta)}{\mu}}. \label{eq:expressionP(E_c|Z_3)_b}
\end{eqnarray}

Considering the case where $\forall i$, $\mu = \lambda_i$, and setting for simplificity $\alpha_{lbert} = k \mu$, we see that 
\begin{equation}\label{eq:rapport_probas}
P(E_{i\neq c}|Z_3)  \,\, = \,\, P(E_{c}|Z_3) \,\, e^{-\dfrac{\alpha}{\mu}} \,\, = \,\, P(E_{c}|Z_3) \,\, e^{-k}  \hskip1truecm \forall \beta \,\, ! 
\end{equation}

\subsubsection{Comments about the results}

\begin{itemize}

\item In this case, we see from Eq.(\ref{eq:rapport_probas}) that the ratio of the different probabilities with respect to $P(E_c|Z_3)$ is independent of $\beta$, which appears only in the expression of $P(E_c|Z_3)$, in Eq.(\ref{eq:expressionP(E_c|Z_3)_b}). When $\alpha = 0$ (case with only the basic constraint), we recover the fact that all the \textbf{possible} probabilities can reach 1, that is, they are all equiprobable \textbf{in their appearance} ! However, in this case with only one more constraint, we see that it is not true anymore and one constraint is different 

\begin{enumerate}
\item if $k>0$, then $P(E_c|Z_3)$ has a different value than the others, and most importantly, even if $P(E_{i\neq c}|Z_3) = e^{-k\beta}\leq 1$ as $0 \leq \beta \leq 1 $, we see that $P(E_c|Z_3) = e^{k(1-\beta)}\geq 1 $ as $k(1-\beta)\geq 0$, that is $P(E_c|Z_3)$ is greater than 1 and so than the others ! It decreases back to 1 when $\beta = 1$ ($ P(E_c|Z_3) = 1$ and the others are 0), or if $\alpha = k=0$ that is if we do not add more informations about the probabilities.

\item if $k<0$, this would have been the contrary. However, the constraints we consider here would not have the same shape : one would be of the form $\sum_i p_i -1$ and the other one of the form $1 - \sum_i p_i$. 

\item if we add more constraints, we would have more parameters to fine-tune, however, we guess that the ratio of probabilities will not depend on the value of $\beta$s.
\end{enumerate}

\item Can we still apply the constraint given in Eq.(\ref{eq:constraints_beta_NN}) ? No, as it would be true only before the (anticiped) update. Doing it would lead to 
\begin{equation} \label{eq:application_probatotales_1}
5e^{-k\beta} + e^{k(1-\beta)} = 1 \,\,\, \Leftrightarrow \,\,\, 5 + e^{k} = e^{k\beta} \,\,\, \Leftrightarrow \,\,\, 5 + e^k - e^{k\beta} = 0
\end{equation}
where we see that $\beta$ can not be equal to 1 : there would have been no unknown probabilities, $P(E_{i\neq c}|Z_3) \rightarrow 0$ and therefore we would not have to applied the constraint in Eq.(\ref{eq:constraints_beta_NN}) and derivatives, as the initial expression setting $\beta = 1$ would have no Ignorance $\bar{h}$. But, even if the minimum would be for 
$k = \dfrac{\ln \beta }{1- \beta }$, due to the fact that $(\beta \neq 0)$ as $e^x>0$ $\forall x$, Eq.(\ref{eq:application_probatotales_1}) will never be fulfilled as $ 5 + e^k - e^{k\beta}>0$ for any value of $k$ and $\beta$. \sk 

Nevertheless, we could argue that in Eq.(\ref{eq:Z1state_constraint_applied}), setting $\mu = \lambda_i$ and after deriving w.r.t $\mu$, the constraint was applied leading to probabilities to be 0 or 1 : It was possible because of these specific values. However, when adding more constraints, this "equilibrium" becomes unbalanced, probabilities are greater than 1 and it is no more possible that a sum of positive values is equal to 1 anymore. In fact, Eq.(\ref{eq:application_probatotales_1}) is wrong in the sense that we have added "expected" probabilities, that is, when $P(E_c|Z_3) = e^{k(1-\beta)}$. Then, because the total sum of probabilities is still 1, the remaining probabilities should be such that $\sum_{i\neq c } P(E_i|Z_3) = 1 - e^{k(1-\beta)}$ ! and not as we did in Eq.(\ref{eq:application_probatotales_1}). However, only if $k<0$, the sum is positive, but $P(E_c|Z_3)$ will not be considered as the most probable, as seen above (except in the framework of negative probabilities, the result would be coherent). \sk 

What we think happens in this case is in fact the following mechanism :  looking only at what becomes the Ignorance with Eq.(\ref{eq:expressionP(E_i|Z_3)_a}), Eq.(\ref{eq:expressionP(E_i|Z_3)_b}) for $P(E_{i\neq c}|Z_3)$ and Eq.(\ref{eq:expressionP(E_c|Z_3)_a}), Eq.(\ref{eq:expressionP(E_c|Z_3)_b}) for $P(E_c|Z_3)$, we see that, setting for simplicity $\lambda_i = \mu$, $\alpha = k \mu$, and assuming that there are $x+1$ propositions where $x$ is the number of ones less probable, 
\begin{eqnarray}
\ln P(E_c|Z_3) = k (1- \beta) & \hskip0.5truecm {}&  P(E_c|Z_3)  = e^{k(1-\beta)} \\
\ln P(E_{i\neq c} |Z_3) = - k \beta & \hskip0.5truecm {}&  P(E_{i \neq c} |Z_3)  = e^{-k\beta} \\
\end{eqnarray}
\begin{eqnarray}
H[..|Z_3]_{postupdate} &=&  \Sigma  + \mu \left[ e^{k(1-\beta)} + xe^{-k\beta} -1 \right] + k\mu(1-\beta)e^{k(1-\beta)} -k\mu x \beta  e^{-k\beta} \\ 
&& - \mu e^{k(1-\beta)} \times k (1- \beta) - \mu x \times e^{-k\beta}  \times (- k \beta ) \\
&=&  \Sigma  + \mu \left[ e^{k(1-\beta)} + xe^{-k\beta} -1 \right] \\ 
&& + k\mu(1-\beta)e^{k(1-\beta)} -k\mu x \beta  e^{-k\beta}  - k \mu (1- \beta) e^{k(1-\beta)}  +  k \mu x \beta e^{-k\beta}   
\end{eqnarray}
and after cancellations, remains only 
\begin{equation}
H[..|Z_3]_{postupdate}  = \Sigma  + \mu \left[ e^{k(1-\beta)} + xe^{-k\beta} -1 \right] 
\end{equation}
But as we have seen, $ \mu \left[ e^{k(1-\beta)} + xe^{-k\beta} -1 \right] $ can not be a constraint and should in fact be replaced by $\alpha_c (P(E_c|Z_3)-1)$. Therefore, one way would be to say that the mechanism which prevents the probabilities to be larger than 1 is, when creating a node in the tree, to set $\mu = 0$ (to clear the previous situation), to set $\alpha_c (P(E_c|Z_3)-1)$ as the constraint wich memorizes what would have happened, and set $\mu(\sum_i p'_i -1)$ as a new constraint taking account of the new situation ($P(E_{i \neq c} |Z_2) \rightarrow p'_i = P(E_{i \neq c}|Z_3)$. \sk 
Thus, the previous worries were in fact ... fine, as it would have been just a mathematical anticipation (maybe a better mathematical description exists, with less worrying aspects as the one where probabilities are not really .. probabilities as here).

\end{itemize}

\subsubsection{A possible interpretation : a network of weighted propositions} \label{sec:interpretation}

Depending on the sign of $k$, and therefore of the value of the Lagrange multiplier $\alpha$, the "most probable" (of probability higher than 1) expected proposition would be either the proposition on which we have more information, or the others. If we consider the case where $k>0$, that could mean that \textit{the more we hear from a proposition, even if at the end its probability will be less than the others ($\beta$-independent !), the more we know about it, and thus, the more we should expect its verification at some point} which seems coherent with what we endure in reality : the more an information is amplified, the more likely we are to come across it in the newspapers, several times and in a short space of time, and the greater our degree of confidence in the veracity of this information. This is of course not correct because we do not know if the information is true, if what we know about it could be coherent, but it is natural to do so, and this can be seen in what we observe here with the probabilities with the constraint $\alpha_c (P(E_c|Z)-\beta)$ : as $\beta$ increases, so is the constraint $\alpha_c$ about it, being another weights we have in the confidence of an information.\sk 
 
If we have information about another proposition, such that $\alpha_d (P(E_d|Z) - \gamma) $ for instance, then another constraint will be added in our brain, and then, depending on the weight of $\alpha_c$ and $\alpha_d$, either $P(E_c|Z)$ or $P(E_d|Z)$ will be more probable. We can imagine than, in a lifetime, Howard will have a lot of proposition to verify (and so a lot of weights to adjust), will hear a lot about some of them, and therefore, some weights will be higher than others, and the total Ignorance will encode all of them as a global quantity but in a really really more complicated expression as the ones we use : all of this in one quantity will account for Howard's state, his knowledge, memories and memory, his assumptions and doubts, ... at a given time; and the processes of minimising his ignorance, as well as his further anticipations using Bayes' formula, are mechanisms that allow him to grasp the complexity of the world around him. But its use will be ineffable to explain in details, in a simpler way, as it could be seen as a neuron network with thousand and thousand of propositions, and relations between them as informations could not be independent !

\section{Conclusion : Memory, Souvenirs, Ignorance, Credence, ...}

In this article, we have followed Howard in his journey to learn, memorize and anticipate, and we have been able to describe it in a global way; using probabilities and Ignorance as defined in \cite{MyIgnorance}. If Howard would have been a particle, cell or anything else, its local description would have also been possible, as in \cite{GiffinCaticha}, \cite{BanavarMaritan} or with the use in the Free Energy Principle as in \cite{FEP}.  \sk

The results we have shown here seem to mathematically account for natural processes that we experience every day in our behaviour when faced with new information(s). Their scope allows for practical but also philosophical applications since we can draw conclusions about how we perceive the world. \sk 

\begin{itemize}
\item As we said previously, it is not because we have a strong credence about a proposition that it will be verified directly. We can however anticipate it, either by minimising the Ignorance directly with the corresponding probability and the weights $\alpha_i$ we have about it (as the values of the probabilities seem to not play a role in the equations), or planning ahead with Bayes'formula. However, because the expected probability can be higher than 1, one can doubt about it, but this is an anticipation process, not something that will be true in reality.

\item Minimising Ignorance seems to be usefull to anticipate the next step, and Bayes'formula to anticipate two steps ahead, the latter being a particular case of the former.

\item We have seen also the philosophical implications of Eq.(\ref{eq:case_of_no_knowledge_consequences}), about Ignorance without Knowledge (see also \cite{PRP}), leading us to the facts that 
\begin{itemize}
\item[$\bullet$] \textit{"I do not have concerns about things I do not know they exist".}\\ \noindent \textit{Je ne m'inqui\`ete pas des choses dont je ne connais pas l'existence.}. 
\item[$\bullet$]  \textit{"I know that I am Ignorant but I do not know about what".}\\ \noindent \textit{J'ignore ce que je ne sais pas.}
\end{itemize}
which are of course obvious, at least for anyone like Howard.

\item Learning and going from state $Z$ to step $Z_0$ to step $Z_1$ ... \textit{etc etc}, at each step/state/node in the tree, constraints and Ignorance were considered but we think we can distinguish between two kind of constraints  : 
\begin{itemize}
\item[$\bullet$] at each node $\alpha_0 [P(E_0|Z) + P(\overline{E_0}|Z) -1]$, with the corresponding Ignorance $- \lambda_0 P(E_0|Z) \ln P(E_0|Z) - \lambda_{\bar{0}}P(\overline{E_0}|Z) \ln P(\overline{E_0}|Z)$. Both do not evolve and describe what was the situation at one particular moment in time : they can be derived and one will find again the same results, however they do not play a role for the other proposition as they have been verified : for instance here, applying the constraint $\alpha_0[P(E_0|Z_0)-1]$ will verify $\alpha_0 [P(E_0|Z) + P(\overline{E_0}|Z) -1] = 0$ and $- \lambda_0 P(E_0|Z) \ln P(E_0|Z) - \lambda_{\bar{0}}P(\overline{E_0}|Z) \ln P(\overline{E_0}|Z) = 0$ . Therefore, their Lagrange multiplier should not be set directly as 0 (meaning Howard would have forgotten them) as they still should appear in the memory (in what we call $\Sigma$ in the previous expression of Ignorance) but as \textbf{"souvenirs/memories"} of configurations of Howard's previous state of Knowledge ! They represent doubts Howard had over time at different states and he can recall them by applying the same processus, even at later states. 

\item[$\bullet$] constraints like $\alpha_0[P(E_0|Z_0)-1] \rightarrow \alpha_0[P(E_0|Z_1)-1]\rightarrow \alpha_0[P(E_0|Z_2)-1] \rightarrow ...$ which "evolve" over the states and assert what are the verified propositions in Howard's memory. They represent knowledge of Howard (relatively to its path in life) he has in memory over time.

\end{itemize}

\item What happens if we forget that a proposition $E_Y$ is true, that we no longer become too sure of its truthfulness ? We still are aware that it is possible and have therefore doubts about it, meaning that the probability/credence is now $P(E_Y|Z_.) = \beta_Y$. As shown before, its whole Ignorance increases by adding another branch in the tree and new weight and terms of Ignorance such that 
\begin{equation}
H[E_Y|Z_.] = \alpha_Y [P(E_Y|Z_.)-\beta_Y] - \lambda_Y P(E_Y|Z_.) \ln P(E_Y|Z_.)
\end{equation}
and then we would still apply naturally what we said in Sec.(\ref{sec:interpretation}) : a loss of certainty is only a new doubt that appears ! 

\end{itemize}

In this paper we have described what could be called a tool, Ignorance, which, like the free energy principle, is derived from Shannon entropy with constraints. We have also described its applications and implications for a person's thinking, learning and remembering process, leading to a mathematical model that could be used to simulate the evolution of a person's memory (see \cite{LeB} for an attempt). Nevertheless, one could go further, especially in the philosophical aspect, by including this model in a much more general framework and draw other conclusions such as the inclusion of Occam's razor and Hume's maxim which seem to be describable by Bayes' formula (see \cite{Cortecs}). However, this work is being saved for later and if you have any thoughts about this work that you would like to share, we would be curious and happy to hear them.

\section{Acknowledgments}

The author would like to express his eternal and deepest gratitude to Abhay, Aurelien, Martin, .. and Deusch, for time and space spend together. Figures were plotted with the help of Geogebra, using Tikz in LateX; and as time passes, knowledge fade, most of the "bad English" has been corrected with the help of DeepL/Translator.



\end{document}